\pdfoutput=1

\documentclass[11pt]{article}

\usepackage[final]{coling}

\usepackage{times}
\usepackage{latexsym}

\usepackage[T1]{fontenc}

\usepackage[utf8]{inputenc}

\usepackage{microtype}

\usepackage{inconsolata}

\usepackage{graphicx}

\usepackage[utf8]{inputenc} 
\usepackage[T1]{fontenc}    
\usepackage{hyperref}       
\usepackage{url}            
\usepackage{booktabs}       
\usepackage{amsfonts}       
\usepackage{nicefrac}       
\usepackage{microtype}      
\usepackage{xcolor}         

\usepackage{graphicx}
\usepackage{multirow}
\usepackage{multicol}
\usepackage{arydshln}
\usepackage{caption}
\usepackage{subcaption}
\usepackage{amsmath}
\usepackage{colortbl}
\usepackage{tikz}
\usetikzlibrary{tikzmark,calc}
\usepackage{enumitem}

\usepackage{pifont}
\newcommand{\cmark}{\ding{51}}%
\newcommand{\xmark}{\ding{55}}%

\newcommand{\grayline}{\arrayrulecolor{gray}\hline\arrayrulecolor{black}}

\newcommand{\red}[1]{{\color{red} #1}}

\newtheorem{definition}{Definition}[section]
\newtheorem{hypothesis}{Hypothesis}[section]

\usepackage{xcolor}
\usepackage[most]{tcolorbox}
\usepackage{float}
\usepackage{xspace}
\tcbset{
  aibox/.style={
    width=474.18663pt,
    top=5pt,
    left=2pt,
    right=2pt,
    bottom=2pt,
    colback=white,
    colframe=black,
    colbacktitle=black,
    enhanced,
    center,
    attach boxed title to top left={yshift=-0.1in,xshift=0.15in},
    boxed title style={boxrule=0pt,colframe=white,},
  }
}
\newtcolorbox{AIbox}[2][]{aibox,title=#2,#1}
\usepackage{booktabs}

\title{How Likely Do LLMs with CoT Mimic Human Reasoning?}

\author{
    Guangsheng Bao\textsuperscript{\rm 1,2,\footnotemark[1]},
    Hongbo Zhang\textsuperscript{\rm 1,2,\footnotemark[1]},  Cunxiang Wang\textsuperscript{\rm 1,2}, Linyi Yang\textsuperscript{\rm 2,3}, and
    Yue Zhang\textsuperscript{\rm 2,3,\footnotemark[2]}
    \\
    \textsuperscript{1} Zhejiang University \\
    \textsuperscript{2} School of Engineering, Westlake University \\
    \textsuperscript{3} Institute of Advanced Technology, Westlake Institute for Advanced Study \\
    \texttt{\{baoguangsheng, zhanghongbo, wangcunxiang, yanglinyi, zhangyue\}@westlake.edu.cn} \\
}

\begin{document}
\maketitle

\renewcommand{\thefootnote}{\fnsymbol{footnote}}
\footnotetext[1]{Equal contribution. \footnotemark[2]Corresponding author.}
\renewcommand{\thefootnote}{\arabic{footnote}}

\begin{abstract}
Chain-of-thought emerges as a promising technique for eliciting reasoning capabilities from Large Language Models (LLMs). However, it does not always improve task performance or accurately represent reasoning processes, leaving unresolved questions about its usage.  In this paper, we diagnose the underlying mechanism by comparing the reasoning process of LLMs with humans, using causal analysis to understand the relationships between the problem instruction, reasoning, and the answer in LLMs.
Our empirical study reveals that LLMs often deviate from the ideal causal chain, resulting in spurious correlations and potential consistency errors (inconsistent reasoning and answers). We also examine various factors influencing the causal structure, finding that in-context learning with examples strengthens it, while post-training techniques like supervised fine-tuning and reinforcement learning on human feedback weaken it. To our surprise, the causal structure cannot be strengthened by enlarging the model size only, urging research on new techniques. We hope that this preliminary study will shed light on understanding and improving the reasoning process in LLM.
\footnote{We release the code and results at \url{https://github.com/StevenZHB/CoT_Causal_Analysis}}

\end{abstract}

\section{Introduction}

\begin{table}[t]
    \centering\small
    \begin{tabular}{m{0.19\linewidth}m{0.71\linewidth}}
        \toprule
        \textbf{Part} & \textbf{Problem Solving Process} \\ 
        \midrule
        \rowcolor{blue!20} \textit{Instruction ($Z$)} & Please act as a math teacher and solve the addition problem ... \newline \# Question: \newline What is the sum of 625126 and 542611? \newline \# Reasoning: \\
        \rowcolor{orange!30} \textit{CoT ($X$)} & Let's add the two numbers digit by digit. \newline 1. The ones place: 6 + 1 = 7 \newline 2. The tens place: 2 + 1 = 3 \newline ... \\
        \rowcolor{green!20}  \textit{Answer ($Y$)} & Therefore, the final computed sum is 1167737. \\
        \bottomrule
    \end{tabular}
    \caption{A simple example for the three parts of a general problem-solving process.}
    \label{tab:intro_case}
\end{table}

Chain of thought (CoT) has become a standard technique for using LLM to solve reasoning tasks \citep{wei2022chain,kojima2022large,wang2023plan}, including complex mathematical reasoning \citep{cobbe2021training,lewkowycz2022solving,imani2023mathprompter} and logical reasoning \citep{liu2023evaluating,xu2023large,pan2023logic}. However, studies show that CoT does not uniformly lead to increased performance \citep{kojima2022large,sprague2023musr} and does not always faithfully represent the true reasoning process in LLM \citep{lanham2023measuring,turpin2023language}, leaving unsolved questions such as \emph{when} and \emph{why} if these issues occur. Intuitively, understanding the mechanism behind and identifying the root cause will be useful in fixing the issues.

\begin{figure*}[t]
    \centering
    \includegraphics[width=0.9\linewidth]{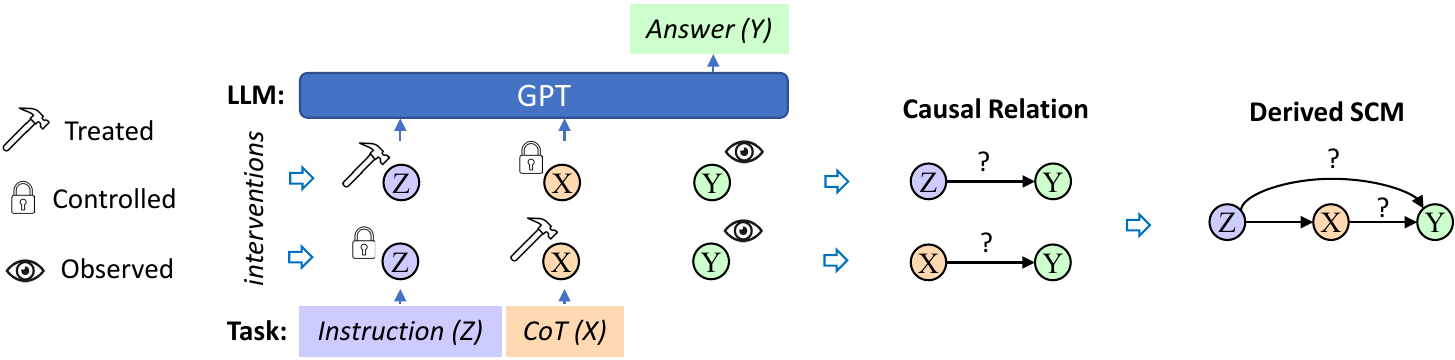}   
    \caption{\emph{Causal analysis}, where we identify an SCM from an LLM-task pair using treatment experiments. For each pair of variables with possible causal relation, we conduct an experiment by injecting an intervention into the treated variable and observe its effect.}
    \label{fig:scm_from_llm}
\end{figure*}

While existing studies have focused on reasoning at the  phenomena level \citep{jin2024impact,harsha2024hardness,yee2024dissociation}, we employ a causal approach, revealing the underlying mechanism of CoT and comparing it with human reasoning. Specifically, without losing generality between tasks and questions, we abstract problem solving into three parts: problem \emph{instruction}, reasoning steps (\emph{CoT}), and conclusion (\emph{answer}), with each part denoted by a random variable that $Z$ for instruction, $X$ for CoT, and $Y$ for answer, as a simple example shown in Table \ref{tab:intro_case}.   We discuss the causal relationship between the three variables for both humans and LLMs. Studies suggest that rational humans follow a \emph{causal chain} when solving complex reasoning problems \citep{cummins1995naive,hegarty2004mechanical,sloman2015causality}, where the instruction causes the reasoning steps and the reasoning steps cause the conclusion. 

\begin{figure*}[t]
    \centering
    \includegraphics[width=1.0\linewidth]{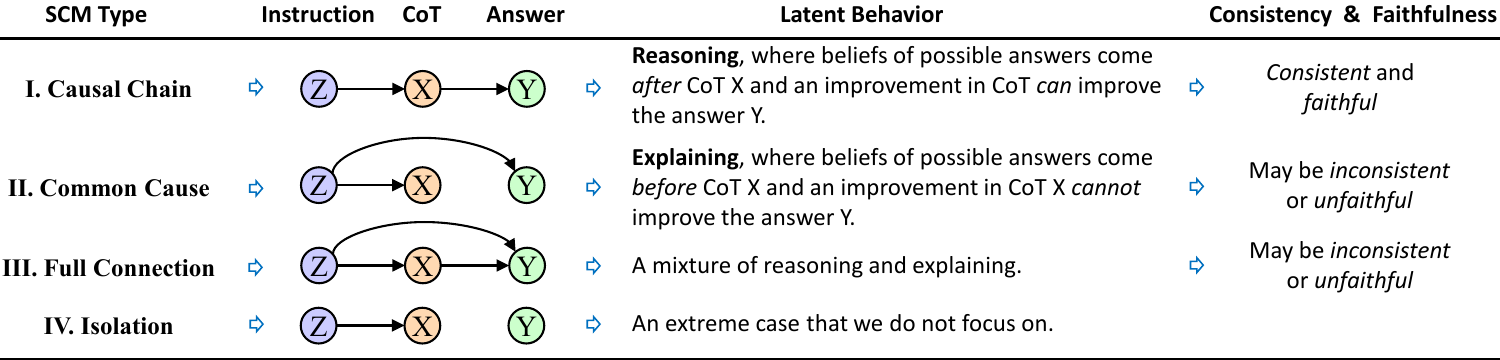}   
    \caption{\emph{Four types of SCM}, where the structure of an SCM reveals its latent behavior, providing explanations on \emph{when} and \emph{why} problems may occur during the reasoning process.}
    \label{fig:causal_relation}
\end{figure*}

For LLMs, we perform \emph{causal analysis} against the three variables by employing interventions \citep{hagmayer2007causal}, assessing the significance of the cause-effect relationship between each pair of variables and assembling a structural causal model (SCM) \citep{pearl2009causality} for each LLM and task pair, as illustrated in Figure \ref{fig:scm_from_llm}.
Specifically, we reveal four types of SCM, including causal chain, common cause, full connection, and isolation (Figure \ref{fig:causal_relation}). Experiments show that a significant portion of LLM-task pairs have the types of common cause (II) and full connection (III), where the model suffers from unexpected spurious correlations between instruction and answer (as the arcs from $Z$ to $Y$ in the SCMs). Empirical evidence indicates that LLMs in these cases may not actually do reasoning (conclude the answer from the CoT) but do explaining (produce the CoT according to the latent belief of the answer). Therefore, reasoning processes can cause an inconsistency error (mismatch between CoT and the answer) and an unfaithful response (discrepancy between CoT and the true reason), as mentioned in the column `{\it consistency \& faithfulness}' in Figure \ref{fig:causal_relation}.


We further investigate various factors that possibly influence the causal structure of implied SCM in six tasks, finding that in-context learning (ICL) strengthens the causal structure while supervised fine-tuning (SFT) \citep{wei2022finetuned} and reinforcement learning on human feedback (RLHF) \citep{ouyang2022training} weakens it. In addition, our investigation of different model sizes reveals that larger language models may not imply stronger type of SCM, suggesting that enlarging model size only may not lead LLMs to ideal human-level reasoning abilities.

Our contributions are mainly twofold:
\begin{enumerate}[label=\arabic*), itemsep=0pt, parsep=4pt]
    \item We discovered \emph{the underlying SCMs of LLMs as essential features}, forecasting their superficial behaviors, such as making consistency errors and producing unfaithful explanations. 
    \item We investigated relevant factors suggesting that \emph{human-level reasoning ability may not be reached by enlarging the model size of LLMs} and popular post-training techniques such as SFT and RLHF actually weaken it.
\end{enumerate}

\section{Related Work}


\paragraph{LLM Reasoning.}
Various reasoning techniques have been proposed to enhance the reasoning ability of LLMs \citep{chu2023survey,reasoning2024survey}. Chain-of-thought (CoT) prompting \citep{wei2022chain}, as an early study elicits reasoning in LLMs, inspires numerous further investigations. Specifically, self-consistency \citep{wang2022self} votes the major decision from multiple reasoning paths, Tree-of-thought \citep{yao2023tree} searches the most confident reasoning path in a tree, and Graph-of-thought \citep{yao2023beyond} represents the thoughts as graph nodes and combines thoughts non-sequentially. Advanced CoT methods, like Faithful CoT \citep{lyu2023faithful} and Constraint CoT \citep{vacareanu2024general}, are further proposed to improve reasoning capabilities.
In this paper, we focus on the very basic chain of thought to understand the underlying mechanism of how LLMs do reasoning, leaving the analysis of advanced methods for the future.

\paragraph{Chain-of-Thought Faithfulness.}
Various recent studies raise concerns about the faithfulness of the reasoning steps generated by CoT prompting. Among them, \citet{turpin2023language} elicits unfaithful reasoning steps using biased model inputs, \citet{paul2024making} finds that the generated answer may not rely on the reasoning steps, \citet{jin2024impact} lengthens the reasoning steps without adding new information but improves the accuracy of the task, \citet{pfau2024let} replaces the reasoning steps with filler tokens to solve algorithmic tasks. 
Furthermore, \citet{lanham2023measuring} proposes a metric to measure the faithfulness of CoT reasoning by intervening the CoT with injected mistakes and altered expressions, but lately \citet{bentham2024chain} doubts about the validity of the metric due to its huge variation under small changes. These studies identify the unfaithfulness of CoT or try to measure it. In this paper, we go further to analyze the latent SCM structures from which we draw connections to various effects, including consistency, faithfulness, and task accuracy.

\paragraph{Causal Reasoning in LLMs.}
Existing studies on the causal reasoning capabilities of LLM mainly focus on variables described in natural language, like benchmark ATOMIC \citep{sap2019atomic}, CLadder \citep{jin2023cladder}, and Corr2Cause \citep{jin2023can}. ATOMIC focuses on the if-then relations of variables like ``{if X pays Y, Y will probably return}''. CLadder requires the identification of variables and their causal relations from the language context prior to inference. Corr2Cause determines the causal structure between variables according to a group of correlational statements.
On multiple causal benchmarks, \citet{kiciman2023causal} finds that LLMs achieve good accuracy and hypothesizes that LLMs can use their collected knowledge to generate causal graphs from natural language, while \citet{zevcevic2023causal} conjectures that a successful causal inference relies on a pre-learned meta-SCM that stores the related causal facts in natural language. 
Unlike these studies, where variables represent targets in the question domain, we investigate the causality between three known variables, instruction, CoT, and answer, which do not represent any question-specific target, but only abstractive components of the chain-of-thought reasoning.

\section{Causal Analysis}
\label{sec:causal_relation_method}

The causal analysis involves three random variables, two hypotheses, and four types of SCM. We explain relevant terminologies such as variables, spurious correlation, and causal relationship in Appendix \ref{app:terminology} and the basic ideas of causal analysis in Appendix \ref{app:scm_confounder}, including the definition of SCM and confounder.

\subsection{Random Variables}

As demonstrated in Table \ref{tab:intro_case}, the problem solving process can be broken down into three random variables, assuming that the question, task, and model remain constant during each experiment.

\textbf{Instruction ($Z$)} typically includes a task description, a few demonstrations, and a question formulation, which guides LLMs in generating a solution or a response. The instruction is restricted by task and question, but the description, demonstrations, and expression can be altered in each experiment.
    
\textbf{CoT ($X$)} signifies the step-by-step reasoning process of an LLM, which results in an answer that is generally considered more precise than a direct answer produced by the LLM. To examine this belief, we distinguish the CoT and the answer as two variables in this paper. 
    
\textbf{Answer ($Y$)} symbolizes the final step of the reasoning process, which answers the question. Ideally, the answer is fully determined by the reasoning steps, which provide complete evidence for the final decision. The answer step is different for each task. For example, ``{\it The correct option is: A}'' for multiple choice tasks, ``{\it The answer is 10.}'' for GSM8K and ``{\it Therefore, the final computed sum is 100.}'' for Addition. Some complete examples are shown in Figure \ref{fig:intro_case} in the Appendix.

\subsection{Identification of SCM}
Intuitively, an autoregressive language model enables the right tokens to depend on all the left tokens, which are represented as a full connection. However, for each specific task, a language model could potentially work in any subgraph of the full SCM. To imply the underlying SCM type of an LLM in a task, we test the causal relations using interventions, focusing on the answer and the relations pointing to it, as illustrated in Figure \ref{fig:scm_from_llm}.

\begin{definition}[Cause-Effect Interventions]
    Suppose that the SCM $\mathcal{G}$ entails a distribution $P_{X,Y}$ with $N_X,N_Y \overset{iid}{\sim} \mathcal{N}(0,1)$. Then we intervene on $X$ to change the distribution of $Y$
    \begin{equation}
    \small
    \begin{split}
        P_Y^{do(X)} = P(Y|do(X)).
    \label{eq:scm-inter}
    \end{split}
    \end{equation}
\end{definition}

Using interventions independent of other variables, we decide whether the treated variable $X$ causes the target variable $Y$. 

\begin{definition}[Average Treatment Effect]
An ATE \citep{rubin1974estimating} represents the effect of an intervention, which compares the distributions of the target variable $Y$ with and without a treatment.
    \begin{equation}
    \small
    \begin{split}
        ATE = E(Y|do(X)) - E(Y).
    \label{eq:ate}
    \end{split}
    \end{equation}
\end{definition}

We assess the significance of the average treatment effects \citep{angrist1995identification} using McNemar's test \citep{mcnemar1947note}.
Specifically, we test two hypotheses: \emph{if the CoT in LLMs causes the answer} and \emph{if the instruction causes the answer}.

\begin{hypothesis}[CoT causes Answer]
    Given a constant Instruction, 
    \begin{equation}
    \small
    \left\{
    \begin{array}{ll}
      H_0: \text{ATE} = 0, & \text{CoT does not cause Answer,} \\
      H_1: \text{ATE} \neq 0, & \text{CoT causes Answer,}
    \end{array}
    \right.
    \end{equation}
    where $\text{ATE} = E(Y|Z,do(X)) - E(Y|Z)$.
\label{hyp:cot_cause_answer}
\end{hypothesis}

\begin{hypothesis}[Instruction causes Answer]
    Given a constant CoT, 
    \begin{equation}
    \small
    \left\{
    \begin{array}{ll}
      H_0: \text{ATE} = 0, & \text{Instruction does not cause Answer,} \\
      H_1: \text{ATE} \neq 0, & \text{Instruction causes Answer,}
    \end{array}
    \right.
    \end{equation}
    where $\text{ATE} = E(Y|X,do(Z)) - E(Y|X)$.
\label{hyp:instruction_cause_answer}
\end{hypothesis}
Based on the statistical significance of the hypotheses, we infer the underlying SCM for LLMs in each task.

Specifically, we test hypothesis \ref{hyp:cot_cause_answer} using two contrasting interventions, golden CoT and random CoT, with expected effects of improving and altering task accuracies, respectively. The \textbf{golden CoT} represents the reasoning steps with golden quality, from which we test if an LLM can conclude the answer correctly. The golden CoT is provided by only four of the datasets. The \textbf{random CoT} is created based on the LLM-generated CoT by injecting mistakes, inspired by \cite{lanham2023measuring}. We randomly replace the numbers in mathematical problems with new digits, and turn the last 1/3 of assertions in logical reasoning problems into negative expressions. More design considerations for CoT interventions are discussed in Appendix \ref{app:cot_intervention}.

Furthermore, we test hypothesis \ref{hyp:instruction_cause_answer} under two conditions, the default CoT and the golden CoT, where the CoTs remain constant during the treatment experiments. The \textbf{default CoT} refers to the CoT generated by the LLM using the default instruction. For each condition, we intervene in the instructions either through random instruction or random bias. The \textbf{random instruction} is different for each task, which is generated by GPT-4 based on the default instruction using a predefined paraphrasing prompt, inspired by the CoT paraphrasing \citep{lanham2023measuring}. We generate a list of alternative instructions for each task and randomly select one for each sample. The design considerations are described in Appendix \ref{app:random_instruction}. The \textbf{random bias} is designed to lead LLMs to wrong answers. We follow \cite{turpin2023language} to inject a bias statement in the instruction like ``{\it I think the correct answer/option is: <the answer>}'', where the answer is either a corrupted golden answer (a digit in the answer is randomly replaced) or a wrong option (a random choice different from the golden selection).

\subsection{SCM Types}
As shown in Figure \ref{fig:causal_relation}, the three random variables can potentially be connected in four directional acyclic graphs (DAGs), each representing a type of SCM.

\textbf{Type I. Causal Chain} is the ideal SCM in which we anticipate LLMs to operate, where the answer $Y$ is entirely determined by CoT $X$. The instruction $Z$ determines the CoT but NOT the answer directly. In other words, the reasoning steps (CoT) fully determine the answer given a question, while other auxiliary information, such as task descriptions and demonstrations, do not directly influence the answer. In such a deterministic relation, we say that CoT \emph{causes} the answer, and CoT is the \emph{only cause} of the answer.

\textbf{Type II. Common Cause} represents an SCM with a confounder instruction $Z$, where the CoT $X$ and the answer $Y$ are not causally connected. We say that the CoT and the answer are conditionally independent given the instruction. Such disconnected CoT and answer are hard to identify from the response only until they produce observable consistency errors.

\textbf{Type III. Full Connection} is the SCM that an autoregressive LLM generally mimics, where the left-to-right causal direction is supported by the causal attention mask \citep{vaswani2017attention} of LLMs. Basically, an LLM is capable of simulating any subgraph of the full SCM. However, the statistical learning of pretraining may not catch the underlying causal structure behind the observation, resulting in superficial behavior-level simulation of human step-by-step reasoning.

\textbf{Type IV. Isolation} denotes an extreme case of SCM, where the answer $Y$ is not influenced by either the instruction $Z$ or CoT $X$. This means that the answer is either determined by the question directly or generated randomly. Due to the complexity of this case, we leave it out of our focus.

\begin{table*}[t]
    \centering\scriptsize
    \renewcommand{\arraystretch}{1.1}
    \begin{tabular}{lccccccc}
        \hline
        \multirow{2}{*}{\bf Experiment} & \multicolumn{7}{c}{\bf GPT-3.5-Turbo}   \\ 
        & {\bf Add.} & {\bf Mult.} & {\bf GSM8K} & {\bf ProofWriter} & {\bf FOLIO} & {\bf LogicQA} & {\bf Avg. $|\text{ATE}|$} \\
        \hline
        Zero-Shot CoT (baseline) & 0.674 & 0.450 & 0.748 & 0.518 & 0.574 & 0.465 & - \\
        \hline
        \multicolumn{8}{c}{\bf Hypothesis Test: If the CoT causes the answer given a constant instruction?} \\
        \grayline
        Controlled (w/ default setting) & 0.782 & 0.454 & 0.742 & 0.520 & 0.588 & 0.480 & - \\
        \multirow{2}{*}{\quad\it Treated (w/ golden CoT)} & \it 0.768 & \it 0.638 & \it 1.000 & \it 0.777 & \it - & \it - & - \\ 
        & \it (-0.014) & \it (+0.184)* & \it (+0.258)* & \it (+0.257)* & \it - & \it - & \it (0.178) \\ 
        \multirow{2}{*}{\quad\it Treated (w/ random CoT)} & \it 0.764 & \it 0.000 & \it 0.018 & \it 0.427 & \it 0.495 & \it 0.440 & - \\
        & \it (-0.018) & \it (-0.454)* & \it (-0.724)* & \it (-0.093)* & \it (-0.093)* & \it (-0.040)* & \it (0.237) \\
        \grayline
        \bf CoT $\overset{?}{\longrightarrow}$ Answer & \bf F & \bf T & \bf T & \bf T & \bf T & \bf T & (0.208) \\
        \hline
        \multicolumn{8}{c}{\bf Hypothesis Test: If the instruction causes the answer given a constant CoT?} \\
        \grayline
        Controlled (w/ default CoT) & 0.782 & 0.454 & 0.742 & 0.520 & 0.588 & 0.480  & - \\
        \multirow{2}{*}{\quad\it Treated (w/ random instruction)} & \it 0.532 & \it 0.488 & \it 0.742 & \it 0.517 & \it 0.578 & \it 0.473 & - \\
        & \it (-0.250)* & \it (+0.034) & \it (+0.000) & \it (-0.003) & \it (-0.010) & \it (-0.007) & \it (0.051) \\
        \multirow{2}{*}{\quad\it Treated (w/ random bias)} & \it 0.228 & \it 0.412 & \it 0.746 & \it 0.503 & \it 0.563 & \it 0.433 & - \\
        & \it (-0.554)* & \it (-0.042) & \it (+0.004) & \it (-0.017)* & \it (-0.025) & \it (-0.047)* & \it (0.115) \\
        \grayline
        Controlled (w/ golden CoT) & 0.768 & 0.642 & 1.000 & 0.782 & - & -  & - \\
        \multirow{2}{*}{\quad\it Treated (w/ random instruction)} & \it 0.510 & \it 0.648 & \it 1.000 & \it 0.612 & \it - & \it - & -  \\
        & \it (-0.258)* & \it (+0.006) & \it (+0.000) & \it (-0.170)* & \it - & \it - & \it (0.109)  \\
        \multirow{2}{*}{\quad\it Treated (w/ random bias)} & \it 0.198 & \it 0.488 & \it 1.000 & \it 0.650 & \it - & \it - & - \\
        & \it (-0.570)* & \it (-0.154)* & \it (+0.000) & \it (-0.132)* & \it - & \it - & \it (0.214) \\
        \grayline
        \bf Instruction $\overset{?}{\longrightarrow}$ Answer & \bf T & \bf T & \bf F & \bf T & \bf F & \bf T & (0.122)  \\
        \hline
        {\bf Implied SCM Type} & {\bf II} & {\bf III} & {\bf I} & {\bf III} & {\bf I} & {\bf III} & - \\
        \hline
    \end{tabular}
    \caption{\emph{Identification of causal structures} in tasks running on \emph{GPT-3.5-Turbo}, where we present task accuracy and ATE. We test the significance of the ATEs from the treated experiments, where the asterisk `*' denotes a statistically significance with a $p$-value $<0.01$ in McNemar's test. The term `{\it default setting}' denotes the default instruction predefined and the default CoT produced by the LLM. It is worth noting that the classification of SCM types is not strictly defined but in the sense of statistical significance.}
    \label{tab:cot_intervention.result_chatgpt}
\end{table*}

\section{Experiments}

\subsection{Experimental Settings}
We evaluate the performance of the model in tasks in terms of \emph{accuracy} by comparing the generated responses with the reference responses. See Appendix \ref{app:prompt_templates} for a detailed description of the prompts used.
 
\paragraph{Models.}
We select multiple open-source and API-based models in experiments, including \emph{ChatGPT}~\citep{openai2022chatgpt}, \emph{GPT-4}~\citep{openai2023gpt4}\footnote{gpt-3.5-turbo-0613 and gpt-4-0613, respectively.}, and open-source models such as the \emph{Llama2} family~\citep{touvron2023llama2} and the \emph{Mistral} series~\citep{jiang2023mistral}. For proprietary models, we use official API calls\footnote{https://openai.com/blog/openai-api}. For open-source models, we employ vLLM~\citep{kwon2023efficient} for local deployment. We use a temperature of 0.0 for all experiments.

\paragraph{Datasets.}
We use six reasoning tasks, primarily mathematical and logical, which are expected to have straightforward, unbiased, and organized solutions. 
The mathematical tasks consisted of basic arithmetic calculations and math word problems, where LLMs demonstrate shortcomings in the former \citep{qian2022limitations} but excel in the latter \citep{wei2022chain}. We create groups of n-digit numbers for elementary-level arithmetic calculations, specifically \emph{Addition} and \emph{Multiplication}. For each of the 6-digit and 9-digit additions and the 2-digit and 3-digit multiplications, 500 samples are generated. Each sample includes a golden step-by-step calculation. For math word problems, we randomly select 500 samples from the \emph{GSM8K} dataset \citep{cobbe2021training}.

We utilize three datasets for the logical tasks. \emph{ProofWriter} \citep{tafjord2020proofwriter} is commonly used for deductive logical reasoning. We randomly select 600 instances from the 5-hop reasoning development set. \emph{FOLIO} \citep{han2022folio} is another dataset for deductive logical reasoning, notable for its expertly crafted content that is more reflective of real-world scenarios. We use all 204 instances from the development set. \emph{LogiQA} \citep{liu2023logiqa} is a dataset drawn from questions in the verbal reasoning exam that require complex logical reasoning abilities. We randomly select 600 entries from the LogiQA 2.0 test set.

\paragraph{Evaluation of Consistency Error.}
We evaluate the consistency between the CoTs and answers. For arithmetic tasks, reasoning steps (CoTs) are extracted and converted into equations using GPT-3.5-turbo prompts (see Appendix \ref{app:prompt_convert}). The generated equations are then compared to the standard (golden) equations to determine the correctness of the generated CoTs. We use the full datasets and report the ratio for each type of error.

For GSM8K and logical tasks, we perform a manual evaluation. A CoT is considered incorrect if it exhibits logical fallacies, contains factual inaccuracies, or fails to deduce the correct answer from the question. We randomly select 200 instances per task and manually examine the generated CoTs by two independent checkers (authors), achieving $96\%$ agreement on 100 overlapped instances.

\subsection{Causal Structures in LLM Tasks}

\textbf{Deriving SCM from treatment experiments.}
Taking \emph{GPT-3.5-turbo} as an example, we illustrate how to interpret the treatment results. As shown in Table~\ref{tab:cot_intervention.result_chatgpt}, by testing the two hypotheses, we obtain qualitative indicators (with a value of `{\it T}' for true or `{\it F}' for false) of causal relationships in each task. Based on these indicators, we imply the SCM.
Specifically, the GPT-3.5-turbo in GSM8K and FOLIO exhibits type I SCM, where answers are NOT significantly affected by interventions in instructions but are significantly affected by interventions in CoTs, suggesting that the LLM in these tasks tends to do real reasoning.

GPT-3.5-turbo in Addition task implies type II SCM, where the answers depend on the instructions but not on the CoTs. Treatment with golden CoTs does not improve accuracy and treatment with random CoTs does not reduce accuracy, suggesting the independence between CoT and answer. In this case, the CoT actually performs explaining instead of reasoning. GPT-3.5-turbo in Multiplication, ProofWriter, and LogiQA implies the full connection, where the latent behavior of CoT is a mixture of explaining and reasoning.

\paragraph{Distribution of SCM types.}
We further collect SCM types for other LLMs such as \emph{GPT-4}, \emph{Llama2-7B-Chat} and \emph{Llama2-70B-Chat}, where the experiments are presented in Table \ref{tab:cot_intervention.result_gpt4}, \ref{tab:cot_intervention.result_llama2_7b}, and \ref{tab:cot_intervention.result_llama2_70b}, respectively, in Appendix \ref{app:detailed_results}. From these experiments, we obtain the distribution of SCM types.

As shown in Table \ref{tab:cot_intervention.implied_scm}, different LLMs suggest different types of SCM. Among them, type III (full connection) is the most common case (10 out of 24 LLM tasks), indicating that most LLMs perform a mixed behavior of reasoning and explaining. 
In smaller Llama2 models, the inferred SCMs are more likely to be type II, III, and IV rather than type I. Although larger GPT-3.5-turbo and GPT-4 show more times of type I, they still have a significant portion in types II and IV.
Consequently, larger LLMs do not necessarily approach the ideal causal chain, suggesting that enlarging the model size only may not lead LLMs to human-level reasoning.

LLMs have different SCM types in different tasks, suggesting their inconsistent ability in different tasks. In this sense, the estimated SCM provides a meaningful indicator of the ability of the model, which can predict the possible mistakes that the model may produce.

\begin{table}[t]
    \centering\scriptsize
    \begin{tabular}{@{}cccccc@{}}
        \hline
        \multirow{2}{*}{\bf SCM} & \multicolumn{2}{c}{\bf Meta Llama2} & \multicolumn{2}{c}{\bf OpenAI GPT} & {\bf \#LLM}  \\         
        & {\bf Chat/7B} & {\bf Chat/70B} & {\bf GPT3.5/175B} & {\bf GPT4/?B} & {\bf -Tasks} \\ 
        \hline
        \multirow{2}{*}{I} & \multirow{2}{*}{-} & \multirow{2}{*}{-} & \multirow{2}{*}{\shortstack{GSM8K \\ FOLIO}} & \multirow{2}{*}{\shortstack{GSM8K \\ Mult.}}  & \multirow{2}{*}{4} \\
        \\
        \hline
        \multirow{3}{*}{II} & \multirow{3}{*}{\shortstack{Addition \\ FOLIO \\ LogiQA}} & \multirow{3}{*}{-} & \multirow{3}{*}{Addition} & \multirow{3}{*}{Addition} & \multirow{3}{*}{5} \\
        \\ \\
        \hline
        \multirow{4}{*}{III} & \multirow{4}{*}{\shortstack{GSM8K \\ ProofW.}} & \multirow{4}{*}{\shortstack{GSM8K \\ ProofW. \\ Addition \\ Mult.}} & \multirow{4}{*}{\shortstack{ LogiQA\\ProofW. \\ Mult. }} & \multirow{4}{*}{ProofW.} & \multirow{4}{*}{10} \\
        \\ \\ \\
        \hline
        \multirow{2}{*}{IV} & \multirow{2}{*}{Mult.} & \multirow{2}{*}{\shortstack{FOLIO \\ LogiQA}} & \multirow{2}{*}{-} &  \multirow{2}{*}{\shortstack{FOLIO \\ LogiQA}} & \multirow{2}{*}{5} \\ 
        \\
        \hline        
    \end{tabular}
    \caption{\emph{Distribution of SCM types}, where the larger models do not necessarily imply stronger SCM types. 
    }
    \label{tab:cot_intervention.implied_scm}
\end{table}

\subsection{When and why do the issues happen?}
We argue that the SCM is an essential feature of an LLM task pair, revealing latent behavior and predicting various superficial problems.

\paragraph{Link to Task Performance.}
Interestingly, the task accuracy of an LLM is not directly related to the type of SCM. When we compare GPT-4 with GPT-3.5-turbo, although GPT-4 achieves a relatively 41\% higher average task accuracy (Table \ref{tab:direct_vs_cot.result} in Appendix \ref{app:direct_vs_cot}), its inferred SCMs do not exhibit more in type I. The type of SCM determines the reasoning process, but not the task accuracy directly.

Consequently, we need different strategies to improve the accuracy of the answers for different types of SCM. For SCM type I, it can be achieved by enhancing the quality of the reasoning steps. However, for SCM type II, because of the conditional independence between the CoT and the answer, it is impossible to achieve better task accuracy by improving the CoT.
These analyses are supported by the treatment experiments with golden CoT, as GSM8K (type I) and Addition (type II) in Table \ref{tab:cot_intervention.result_chatgpt} shows. The golden CoTs (treated w/ golden CoT) improves the task accuracy of GSM8K from $0.742$ to $1.000$ ($+0.258$), but does not improve the task accuracy of Addition.

For SCM type III, it is also possible to improve accuracy by improving the reasoning steps, but there is no guarantee due to the unknown portion of reasoning and explaining underlying the CoT. Specifically, as the Multiplication and ProofWriter in the table show, the treatment with golden CoT improves the task accuracy of Multiplication from $0.454$ to $0.638$ ($+0.184$) and ProofWriter from $0.520$ to $0.777$ ($+0.257$), where improvements are made, but accuracies are still far from perfect $1$.

\paragraph{Link to Faithfulness.}
Given an SCM type, we can predict the faithfulness of the LLM responses. For type I, the LLM tends to produce faithful reasons, while for type II and III, the LLM may produce unfaithful explanations because of the confounder between the CoT and the answer. These forecasts are confirmed by the significant ATEs under random bias treatment, as shown on Addition, Multiplication, ProofWriter, and LogiQA in Table \ref{tab:cot_intervention.result_chatgpt}. The bias changes the beliefs of the answer before CoTs and the answers. As a result, even with constant CoTs (either the default CoT or golden CoT), the answers change into incorrect ones, demonstrating an unfaithful representation of CoT to the real reasoning behind the latent belief.

In practice, none of the LLMs and tasks performs pure reasoning or explaining, but is somehow a mixture of them (as supported by the insignificant but non-zero ATE values in Table \ref{tab:cot_intervention.result_chatgpt}). Therefore, unfaithful responses generally occur in all LLMs and tasks to some extent.

\begin{table}[t]
    \centering\scriptsize
    \begin{tabular}{@{}c@{\hspace{10pt}}l@{\hspace{10pt}}c@{\hspace{5pt}}c@{\hspace{10pt}}c@{\hspace{5pt}}c@{}}
        \toprule
        \multirow{2}{*}{\textbf{SCM}} & \multirow{2}{*}{\textbf{Behavior}} & \multicolumn{2}{c}{\textbf{GPT-3.5-Turbo}} & \multicolumn{2}{c}{\textbf{GPT-4}} \\
        \cmidrule(lr){3-4} \cmidrule(lr){5-6}
        & & \textbf{Task} & \textbf{Error Rate} & \textbf{Task} & \textbf{Error Rate} \\
        \midrule
        \multirow{2}{*}{I} & \multirow{2}{*}{Reasoning} & GSM8K & \colorbox{red!1}{0.000} & GSM8K & \colorbox{red!1}{0.000} \\
        & & FOLIO & \colorbox{red!12}{0.125} & Multi & \colorbox{red!18}{0.178} \\
        \midrule
        II & Explaining & Addition & \colorbox{red!65}{0.648} & Addition & \colorbox{red!74}{0.744} \\
        \midrule
        \multirow{3}{*}{III} & \multirow{3}{*}{Mixture} & LogiQA & \colorbox{red!12}{0.125} & & \\
        & & ProofWriter & \colorbox{red!28}{0.280} & ProofWriter & \colorbox{red!6}{0.060} \\
        & & Multi & \colorbox{red!44}{0.444} & & \\
        \bottomrule
    \end{tabular}
    \caption{\emph{Correspondence between SCM type and consistency error}, where the intensity of \colorbox{red!75!white}{red} denotes the severity of the consistency errors, with darker colors indicating higher rates. Overall, the reasoning behavior corresponds to the least error rate, while the explaining behavior the most and the mixture behavior the middle. }
    \label{tab:reasoning_explaining.consistency}
\end{table}

\begin{table*}[h]
    \centering\scriptsize
    \renewcommand{\arraystretch}{1.4}
    \begin{tabular}{l|c@{\hspace{5pt}}c@{\hspace{5pt}}c@{\hspace{5pt}}c@{\hspace{5pt}}c@{\hspace{5pt}}c@{\hspace{5pt}}c|c@{\hspace{5pt}}c@{\hspace{5pt}}c@{\hspace{5pt}}c@{\hspace{5pt}}c@{\hspace{5pt}}c@{\hspace{5pt}}c|c}
        \hline
        \multirow{2}{*}{\bf ICL} & \multicolumn{7}{c}{\bf $|\text{ATE}|$ on `CoT $\overset{?}{\longrightarrow}$ Answer' ($\uparrow$)} & \multicolumn{7}{c|}{\bf $|\text{ATE}|$ on `Instruction $\overset{?}{\longrightarrow}$ Answer' ($\downarrow$)} & \bf Task \\ 
        & \bf Add. & \bf Mul. & \bf GSM. & \bf Pro. & \bf FOL. & \bf LQA. & \bf Avg. & \bf Add. & \bf Mul. & \bf GSM. & \bf Pro. & \bf FOL. & \bf LQA. & \bf Avg. & \bf Accuracy \\
        \hline
        0-shot & 0.016$_F$ & 0.319$_T$ & 0.491$_T$ & 0.175$_T$ & 0.093$_T$ & 0.040$_T$ & 0.208 & 0.408$_T$ & 0.059$_T$ & 0.001$_F$ & \bf 0.081$_T$ & \bf 0.018$_F$ & 0.027$_T$ & 0.122 & 0.572 \\
        2-shot & \bf 0.095$_T$ & 0.338$_T$ & 0.497$_T$ & 0.254$_T$ & \bf 0.265$_T$ & \bf 0.043$_T$ & \bf 0.251 & 0.104$_T$ & 0.035$_T$ & \bf 0.000$_F$ & 0.117$_T$ & 0.073$_T$ & 0.006$_F$ & 0.059 & \bf 0.598 \\
        4-shot & 0.014$_F$ & 0.336$_T$ & \bf 0.498$_T$ & 0.251$_T$ & 0.235$_T$ & 0.022$_F$ & 0.227 & \bf 0.044$_T$ & \bf 0.034$_T$ & \bf 0.000$_F$ & 0.118$_T$ & 0.064$_T$ & \bf 0.002$_F$ & \bf 0.049 & 0.580 \\
        8-shot & 0.026$_T$ & \bf 0.342$_T$ & 0.497$_T$ & \bf 0.267$_T$ & 0.186$_T$ & 0.035$_T$ & 0.229 & 0.093$_T$ & 0.044$_T$ & \bf 0.000$_F$ & 0.181$_T$ & 0.089$_T$ & 0.007$_F$ & 0.065 & 0.592 \\
        \hline
    \end{tabular}
    \caption{\emph{The impact of ICL} on causal relationships tested on \emph{GPT-3.5-Turbo}, where the best $|\text{ATE}|$ and task accuracy are marked in bold. The `{\it T/F}' indicates the statistical significance of the causal relation. The detailed outcomes of the 0/2/4/8-shot can be found in Table \ref{tab:cot_intervention.result_chatgpt} and Table \ref{tab:cot_intervention.result_chatgpt_2shot}, \ref{tab:cot_intervention.result_chatgpt_4shot},  \ref{tab:cot_intervention.result_chatgpt_8shot} in Appendix \ref{app:detailed_results}, respectively.}
    \label{tab:cot_intervention.implied_scm_nshot}
\end{table*}

\paragraph{Link to Consistency Error.}
We evaluate the consistency of CoTs and answers in six tasks, finding that incorrect CoTs may be followed by correct answers and vice versa. As Table \ref{tab:direct_vs_cot.confusion} in the Appendix shows, on five of the six tasks, LLMs produce consistency errors, particularly on simple arithmetic problems like Addition and Multiplication. For example, more than 60\% incorrect CoTs lead to correct answers in Addition, and a larger model such as GPT-4 shows even more discrepancy of 74\% (Appendix \ref{app:cot_inconsistency}).

Intuitively, the reasoning behavior tends to produce consistent responses because the answers are concluded from the reasoning steps, while the explaining behavior may produce inconsistent CoTs and answers because they stochastically depend on the same latent beliefs. We examine the types of SCM and the consistency error rates as shown in Table \ref{tab:reasoning_explaining.consistency}. 
The results indicate that the tasks with type I SCM generally have lower error rates than the tasks with type II SCM, suggesting that the former makes fewer consistency errors because of the strong causal connection between the CoT and the answer. 
The untypical SCM type III is expected to sit between types I and II because of the mixed behavior, which is partially supported by the error rates (larger than or equal to type I generally but smaller than type II).

\subsection{How to fix the issues?}
We investigate possible factors that influence causal structures, including ICL, SFT, and RLHF.

\begin{table*}[h]
    \centering\scriptsize
    \renewcommand{\arraystretch}{1.4}
    \begin{tabular}{l|c@{\hspace{5pt}}c@{\hspace{5pt}}c@{\hspace{5pt}}c@{\hspace{5pt}}c@{\hspace{5pt}}c@{\hspace{5pt}}c|c@{\hspace{5pt}}c@{\hspace{5pt}}c@{\hspace{5pt}}c@{\hspace{5pt}}c@{\hspace{5pt}}c@{\hspace{5pt}}c|c@{}}
        \hline
        \multirow{2}{*}{\bf Model} & \multicolumn{7}{c}{\bf $|\text{ATE}|$ on `CoT $\overset{?}{\longrightarrow}$ Answer' ($\uparrow$)}  & \multicolumn{7}{c|}{\bf $|\text{ATE}|$ on `Instruction $\overset{?}{\longrightarrow}$ Answer' ($\downarrow$)} & \bf Task \\ 
        & \bf Add. & \bf Mul. & \bf GSM. & \bf Pro. & \bf FOL. & \bf LQA. & \bf Avg. & \bf Add. & \bf Mul. & \bf GSM. & \bf Pro. & \bf FOL. & \bf LQA. & \bf Avg. & \bf Accuracy \\
        \hline
        Base & 0.006$_F$ & \bf 0.164$_T$ & \bf 0.496$_T$ & \bf 0.254$_T$ & 0.034$_F$ & 0.040$_T$ & \bf 0.204 & \bf 0.006$_F$ & 0.040$_T$ & \bf 0.012$_T$ & 0.411$_T$ & 0.089$_T$ & \bf 0.001$_F$ & 0.111 & \bf 0.313 \\
        SFT & \bf 0.008$_F$ & 0.091$_T$ & 0.490$_T$ & 0.191$_T$ & \bf 0.054$_F$ & 0.023$_F$ & 0.179 & 0.021$_T$ & 0.072$_T$ & 0.098$_T$ & 0.381$_T$ & 0.086$_T$ & 0.200$_T$ & 0.156 & 0.300  \\
        DPO & 0.006$_F$ & 0.026$_T$ & 0.417$_T$ & 0.119$_T$ & 0.049$_F$ & \bf 0.058$_T$ & 0.138 & 0.022$_T$ & \bf 0.005$_F$ & 0.092$_T$ & \bf 0.114$_T$ & \bf 0.042$_F$ & 0.118$_T$ & \bf 0.066 & 0.275 \\
        \hline
    \end{tabular}
    \caption{\emph{The impact of SFT/RLHF} on causal relationships based on \emph{Mistral-7B}, where SFT primarily enhances the causal connection between the instruction and answer, and DPO diminishes this relationship. The detailed outcomes of the Base, SFT, and DPO models can be found in Table \ref{tab:cot_intervention.result_mistral_base}, \ref{tab:cot_intervention.result_mistral_sft}, and \ref{tab:cot_intervention.result_mistral_dpo} in Appendix \ref{app:detailed_results}, respectively.}
    \label{tab:cot_intervention.implied_scm_mistral}
\end{table*}

\paragraph{Impact of In-Context Learning.}
In-context learning (ICL) is commonly employed to elicit expected behaviors in LLM \citep{brown2020language}, which is also used to trigger CoT to mimic human step-by-step reasoning \citep{wei2022chain}. We assess the impact of ICL on causal relationships in our context, with randomly chosen examples as ICL demonstrations. We carry out treatment experiments with varying numbers of demonstrations, as shown in Table \ref{tab:cot_intervention.implied_scm_nshot}.

The results reveal that, compared to zero-shot, ICL demonstrations improve causal relationships and enhance task accuracies. Specifically, ICL generally reduces $|\text{ATE}|$ in `Instruction $\rightarrow$ Answer', but enhances $|\text{ATE}|$ in `CoT $\rightarrow$ Answer' as indicated by the Avg. |ATE|, resulting in improved causal relationships.

\paragraph{Impact of SFT and RLHF.}
Supervised fine-tuning (SFT) enables LLMs to follow human instructions \citep{wei2022finetuned}, while reinforcement learning from human feedback (RLHF) aligns LLMs with human preferences \citep{christiano2017deep,ouyang2022training}. However, recent studies suggest that SFT and RLHF may induce hallucinations \citep{schulman2023reinforcement,yang2023alignment}. We hypothesize that they may also affect the causal structures.

We validate our hypothesis by performing causal analysis on three models: a foundation model Mistral-7B-Base \footnote{https://huggingface.co/mistralai/Mistral-7B-v0.1} \citep{jiang2023mistral}, an instruct-tuned model Mistral-7B-SFT \footnote{https://huggingface.co/HuggingFaceH4/zephyr-7b-sft-beta}, and an RLHF-tuned model Mistral-7B-DPO \footnote{https://huggingface.co/HuggingFaceH4/zephyr-7b-beta} \citep{tunstall2023zephyr}. Since the base model cannot follow instructions, we elicit the question-answering behavior using ICL with four demonstrations, and similarly, we use the same demonstrations for the SFT and DPO models.

As Table \ref{tab:cot_intervention.implied_scm_mistral} shows, SFT generally weakens the causal structure, with a smaller Avg. $|\text{ATE}|$ on `CoT $\rightarrow$ Answer' but a larger Avg. $|\text{ATE}|$ on `Instruction $\rightarrow$ Answer', indicating that SFT introduces spurious features into the model, thereby causing hallucinations. In contrast, DPO reduces spurious features by weakening the link between the instruction and the answer and reducing the Avg. $|\text{ATE}|$ from $0.111$ to $0.066$. The findings align with the human preference to separate the answers from irrelevant spurious features \cite{ouyang2022training}.
However, DPO also weakens the causal connection between CoTs and answers (lower the Avg. $|\text{ATE}|$ to $0.138$), suggesting a negative side effect of DPO.



\section{Conclusion}
We conducted causal analyses on LLMs with CoT, revealing the underlying SCM structures, which serve as essential features that can be used to predict the latent behaviors, and further consistency and faithfulness of CoT. Analyses of the relevant factors show that model size has a significant influence on the causal structure, but larger models do not necessarily lead to better SCMs. Popular techniques like ICL, SFT, and RLHF affect causal structures, with ICL strengthening them while SFT and RLHF weakening them.

Our findings underscore the need for further research into effective LLM techniques to strengthen causal structures, with the goal of achieving human-level reasoning ability.

\newpage

\section*{Acknowledgments}
We would like to thank the anonymous reviewers for their valuable feedback. This work is funded by the National Natural Science Foundation of China Key Program (Grant No. 62336006) and the Pioneer and “Leading Goose” R\&D Program of Zhejiang (Grant No. 2022SDXHDX0003).

\section*{Limitations}
This study focuses primarily on the analysis of existing model and LLM techniques on their impact on the underlying causal structures, leaving the exploration of new techniques to improve the causal structure in the future. We also focus on the currently popular Generative Pre-Training (GPT) \citep{radford2018improving} language models, setting aside other models such as the Bidirectional Encoder Representations from Transformers (BERT) \citep{devlin2018bert} and the General Language Model (GLM) \citep{du2022glm} for future exploration. This is due to the potentially more intricate causal structures in these models, stemming from their blank-infilling training objective. Furthermore, our research predominantly deals with standard mathematical and logical reasoning, not including areas like common sense and symbolic reasoning within its scope.

\section*{Ethical and Broader Impact}
The study offers a framework for understanding the decision-making process and reasoning of LLM, which could contribute to greater transparency and accountability in AI systems. It underscores the fact that LLMs can be swayed by unrelated contexts, resulting in biased outcomes. The study implies that the typical techniques employed in LLM might not necessarily improve its reasoning abilities. This could impact the way we train and educate AI models in the future.

\bibliography{custom}

\newpage

\appendix

\section{Terminology}
\label{app:terminology}

We provide a brief overview of the concepts about causal analysis, which are mainly from \citet{peters2017elements}.

\textbf{Variables}: Variables are quantities, characteristics, or properties that can be measured or observed. They can change or vary and typically fall into two categories: dependent variables, which are being tested and measured in an experiment, and independent variables, which are manipulated or controlled in an experiment. In this paper, the three variables refer to the three parts involved in the CoT reasoning.

\textbf{Spurious Correlation}: Spurious correlation refers to a mathematical relationship in which two or more variables are not causally related to each other, yet it may be wrongly inferred that they are, due to either coincidence or the presence of a certain third, unseen factor (referred to as a "confounding factor"). The statistical training of LLMs introduce spurious correlations because without treatment experiments, the spurious correlation and causal correlation cannot be inferred from the training data alone.

\textbf{Causal Relation}: A causal relation between two variables exists if the occurrence of the first causes the other. The first variable is called the cause, and the second variable is called the effect. A causal relationship is often established by methodically manipulating the cause and observing the effect. In this paper, we use systematic treatment experiments to detect the significance of the causal relations.

\textbf{Causal Structure Model (SCM)}: A causal structure model (also known as a causal model or structural causal model) is a conceptual model that describes the causal mechanisms of a system. The model is usually formalized as directed acyclic graph (DAG), where the nodes represent the variables and the edges represent causal relationships between the variables. This model helps in understanding how a system works and predicting the effects of interventions.

\textbf{Causal Chain}: A sequence of variables, each one triggering the next. In a causal chain, a variable occurs due to some cause, which itself is the effect of some other cause, and so on. It helps to trace the root cause of any variable. In the context of CoT reasoning, the expected causal relations between three variables form a causal chain: the task description (instruction, variable Z) decides the CoT reasoning steps (variable X), and the reasoning steps decide the final answer (variable Y). Such a causal chain exists in the correct logical and mathematical reasoning process.

\section{Causal Analysis}
\label{app:scm_confounder}

Causal analysis is a method that is used to identify and understand the causes and effects of different actions, situations, or decisions. It involves examining the reasons or causes behind a certain occurrence and the outcomes that may arise from it \citep{heise1975causal, imbens2015causal,feder2022causal}. Causal models in different structures may induce the same observational distribution but different intervention distributions \citep{peters2017elements}. Interventions thus can be used to differentiate among the potential causal structures that are compatible with an observation \citep{hagmayer2007causal, pearl2009causality}. We study CoT by its causal relationship to the model decisions, using interventions to test the significance of the cause-effect relations between CoTs/instructions and answers.

The basic concepts of SCM and confounder are described as follows.

\begin{definition}[Structural Causal Model]
    A simple SCM $\mathcal{G}$ with a graph $X \rightarrow Y$ consists of two random variables $X$ and $Y$, following assignments
    \begin{equation}
    \small
    \begin{split}
        X&:=N_X, \\
        Y&:=f_Y(X,N_Y),
    \label{eq:scm-def}
    \end{split}
    \end{equation}
    where the noise $N_Y$ is independent of $N_X$.

    We refer to the random variables $X$ as the \textbf{cause} and $Y$ as the \textbf{effect}. The graph $X \rightarrow Y$ indicates that $X$ is a direct cause of $Y$.
\end{definition}

\begin{definition}[Confounder of Variables]
    An SCM $\mathcal{G}$ with a graph consists of three random variables $X$, $Y$, and $Z$, following assignments
    \begin{equation}
    \small
    \begin{split}
        X&:=f_X(Z,N_{XZ}), \\
        Y&:=f_Y(Z,N_{YZ}),
    \label{eq:confounder-def}
    \end{split}
    \end{equation}
    where the noise $N_{YZ}$ is independent of $N_{XZ}$.

    We refer to the random variable $Z$ as the \textbf{confounder} of the random variables $X$ and $Y$.
\end{definition}

\subsection{Design of CoT Interventions}
\label{app:cot_intervention}
The golden CoT is designed to enhance accuracy by offering clear hints toward the correct solution, whereas the random CoT is likely to reduce accuracy due to the inclusion of distorted information (such as incorrect numbers for math tasks or misleading statements for logic problems), which can misguide the reasoning process toward an incorrect conclusion.

The golden CoT does not leave much room for considering different options. On the other hand, random CoT presents various alternatives. We experiment with several techniques to disrupt the CoT steps. For example, we negate one third of the CoT statements at different points -- beginning, middle, and end -- and observe that altering the beginning or middle parts did not significantly affect the outcomes. Consequently, we focus on the end part.

We also test other disturbance strategies for the CoT, like mixing up the names of entities or altering the sequence of reasoning steps. These methods typically cause the model to initiate a new line of reasoning, thus disregarding the manipulated CoT. In contrast, our chosen method of intervention successfully disrupts the CoT while still keeping the model's dependence on the given reasoning pathway intact.

\subsection{Design of Random Instruction}
\label{app:random_instruction}
To intervene in the instructions within the prompt, we adhere to the following principles: 1) There is a controllable variable to adjust the direction of the intervention. 2) There is a clear distinction between the pre- and post-intervention expressions. 3) The intervened instructions should still guide the model in derive the answer through CoT.
Based on these principles, we instruct the GPT-4 to paraphrase the original instructions and control the target distribution of the paraphrase through predefined arbitrary roles (professions). As an example, we present the original and intervened instructions for GSM8K in Table~\ref{tab:intervene_instruction_demo}.
\begin{table}[h!]
    \centering\small
    \begin{tabular}{m{0.15\linewidth}m{0.75\linewidth}}
        \toprule
        \textbf{Role} & \textbf{Instruction} \\ 
        \midrule
        \textit{Original} & Please act as a math teacher and solve the math problem step by step. \\ \midrule
         & \textit{Intervened Instructions:} \\ \addlinespace[0.5em]
        \textit{Chef} & Imagine you are a chef in a bustling kitchen, and you need to tackle this math problem as if it were a recipe. Break down the solution into clear, step-by-step instructions. \\ \addlinespace[0.5em]
        \textit{Detective} & Imagine you are a detective unraveling a mystery. Solve the problem meticulously, step by step, as you would piece together clues in an investigation. \\ \addlinespace[0.5em]
        \textit{Judge} & I need you to take on the role of a judge and adjudicate the math problem, providing a detailed step-by-step resolution. \\ \addlinespace[0.5em]
        \textit{Artist} & Imagine you are an artist, and approach solving the math problem with creativity and flair, breaking it down into steps. \\ 
        \bottomrule
    \end{tabular}
    \caption{GSM8K Original and Intervened Instructions}
    \label{tab:intervene_instruction_demo}
\end{table}

\section{Experimental Settings}
\subsection{Prompts and Templates}
\label{app:prompt_templates}

We use general prompts and templates for experiments with \emph{Direct} answering and zero-shot \emph{CoT}. To facilitate answer matching, we instructed the model to respond using a template format, based on the approach of prompt modifications by \citet{pan2023logic}. The prompts for each task are as follows.

\noindent
\begin{minipage}{1.0\linewidth}
\begin{tabular}{p{18.5em}}
    \textbf{\small Addition (Direct):} \\
    \midrule
    \parbox{\linewidth}{\small 
        Please act as a math teacher and solve the math problem. Please directly answer with the format "The answer is <<answer>>" without any other information. \\

        What is the sum of \{\{number1\}\} and \{\{number2\}\}?
    }  \\
    \vspace{4pt}
    \textbf{\small Addition (CoT):} \\
    \midrule
    \parbox{\linewidth}{\small 
        Please act as a math teacher and solve the addition problem in the given template. \\
        \#\#\#\# \\
        \# Question: \\
        What is the sum of <<number1>> and <<number2>>? \\
        \# Reasoning: \\
        Let's add the two numbers digit by digit. \\
        1. The ones place: <<digit1>> \\
        2. The tens place: <<digit2>> \\
        <<other digits>> \\
        Answer: \\
        Therefore, the final computed sum is <<answer>>. \\
        \#\#\#\# \\
        \# Question: \\
        What is the sum of \{\{number1\}\} and \{\{number2\}\}? \\
        \# Reasoning:
    }  \\
    \vspace{4pt}
    \textbf{\small Multiplication (Direct):} \\
    \midrule
    \parbox{\linewidth}{\small 
        Please act as a math teacher and solve the math problem. Please directly answer with the format "The answer is <<answer>>" without any other information. \\

        What is the product of \{\{number1\}\} and \{\{number2\}\}?
    }  \\
    \vspace{4pt}
    \textbf{\small Multiplication (CoT):} \\
    \midrule
    \parbox{\linewidth}{\small 
        Please act as a math teacher and solve the product problem in the given template. \\
        \#\#\#\# \\
        \# Question: \\
        What is the product of <<number1 such as 27>> and <<number2 such as 153>>? \\
        \# Reasoning: \\
        Let's think step by step. <<number2 153 has three digits, so that we can reason in three steps.>> \\
        1. <<Multiply number1 27 by the ones place digit 3 of number2 153>> \\
        2. <<Multiply number1 27 by the tens place digit 50 of number2 153>> \\
        <<other digits of number2 if it has>> \\
        Answer: \\
        Now, sum all the step results: <<sum of the results>>. \\
        So, the final computed product is <<answer>>. \\
        \#\#\#\# \\
        \# Question: \\
        What is the product of \{\{number1\}\} and \{\{number2\}\}? \\
        \# Reasoning:
    }  \\
\end{tabular}
\end{minipage}

\noindent
\begin{minipage}{1.0\linewidth}
\begin{tabular}{p{17.5em}}
    \textbf{\small GSM8K (Direct):} \\
    \midrule
    \parbox{\linewidth}{\small 
        Please act as a math teacher and solve the math problem. Please directly answer with the format "The answer is <<answer>>" without any other information. \\

        \{\{question\}\}
    }  \\
    \vspace{4pt}
    \textbf{\small GSM8K (CoT):} \\
    \midrule
    \parbox{\linewidth}{\small 
        Please act as a math teacher and solve the math problem step by step. \\
        \# Question: \\
        \{\{question\}\} \\
        \# Reasoning: \\
        Let's think step by step.
    }  \\
    \vspace{4pt}
    \textbf{\small ProofWriter/FOLIO/LogiQA (Direct):} \\
    \midrule
    \parbox{\linewidth}{\small 
        Your goal is to solve the logical reasoning problem. Given a context and a question, directly answer with the format "The correct option is: A/B/C" without any other information. \\
        \#\#\#\# \\
        \# Context: \\
        \{\{context\}\} \\
        
        \# Question: \\
        \{\{question\}\} \\
        \# Options: \\
        \{\{options\}\} \\
        
        \# Instruction: 
        \#\# Answer:
    }  \\
    \vspace{4pt}
    \textbf{\small ProofWriter/FOLIO/LogiQA (CoT):} \\
    \midrule
    \parbox{\linewidth}{\small 
        Please act as a math teacher and reason step by step to solve the logical reasoning problem. Given a context and a question, explain your reasoning process and give the answer with the format "The correct option is: A/B/C". \\
        \#\#\#\# \\
        \# Context: \\
        \{\{context\}\} \\
        
        \# Question: \\
        \{\{question\}\} \\
        \# Options: \\
        \{\{options\}\} \\
        
        \# Instruction: \\
        \#\# Reasoning: \\
    }  \\
\end{tabular}

\noindent For LogiQA, the correct option is: A/B/C/D.

\vspace{10pt}

\end{minipage}

Readers may wonder why the CoT prompts for Addition and Multiplication look so different from others and why we do not just simply use ``let's think step by step''. In practice, the instruction ``let's think step by step'' does not consistently trigger a step-by-step reasoning process for Addition and Multiplication tasks on both GPT-3.5-turbo and GPT-4. Therefore, we tested 20 samples each on GPT-3.5-turbo and GPT-4 to derive the most common templates for regulating the responses.

\subsection{Normalize Reasoning Steps}
\label{app:prompt_convert}
We use GPT-3.5-turbo to normalize the generated reasoning steps for Addition and Multiplication. As the following prompts show, the conversion from verbal reasoning steps to formal expressions is fairly straightforward. We manually reviewed 50 random samples for each task and observed accurate conversions.

\noindent
\begin{minipage}{1.0\linewidth}
\begin{tabular}{p{18.5em}}
    \textbf{\small Addition} \\
    \midrule
    \parbox{\linewidth}{\small 
        Please convert the natural language described reasoning steps into formal expressions as the examples. Please put the carry 1 at the last of the addition of each step. \\
        \#\#\#\# \\
        \# Reasoning Steps: \\
        Let's add the two numbers digit by digit. \\
        1. The ones place: 0 + 0 = 0 \\
        2. The tens place: 2 + 9 = 11 (carry over the 1) \\
        3. The hundreds place: 7 + 8 + 1 = 16 (carry over the 1) \\
        4. The millions place: 1 + 2 + 1 = 4 \\
        \# Formal Expressions: \\
        1. 0 + 0 = 0 \\
        2. 2 + 9 = 11 \\
        3. 7 + 8 + 1 (carry) = 16 \\
        4. 1 + 2 + 1 (carry) = 4 \\
        \#\#\#\# \\
        … (other examples) \\
        \#\#\#\# \\
        \# Reasoning Steps: \\
        {{reason}} \\
        \# Formal Expressions: \\
    }  \\
\end{tabular}
\end{minipage}

\noindent
\begin{minipage}{1.0\linewidth}
\begin{tabular}{p{18.5em}}
    \textbf{\small Multiplication} \\
    \midrule
    \parbox{\linewidth}{\small 
        Please convert the natural language described reasoning steps into formal expressions as the examples. \\
        \#\#\#\# \\
        \# Reasoning Steps: \\
        Let's think step by step. 305 has three digits, so that we can reason in three steps. \\
        1. Multiply 487 by the ones place digit 5 of 305. The result is 2435. \\
        2. Multiply 487 by the tens place digit (0*10) of 305. The result is 0. \\
        3. Multiply 487 by the hundreds place digit (3*100) of 305. The result is 146100. \\
        \# Formal Expressions: \\
        1. 487 * 5 = 2435 \\
        2. 487 * 0 = 0 \\
        3. 487 * 300 = 146100 \\
        \#\#\#\# \\
        … (other examples) \\
        \#\#\#\# \\
        \# Reasoning Steps: \\
        {{reason}} \\
        \# Formal Expressions: \\
    }  \\
\end{tabular}
\end{minipage}

\begin{table*}[h!]
    \centering\tiny
    \begin{tabular}{ccccccccccc}
        \hline
        \multirow{2}{*}{\bf LLM} & \multirow{2}{*}{\bf Method} & \multicolumn{2}{c}{\bf Addition} & \multicolumn{2}{c}{\bf Multiplication} & \multirow{2}{*}{\bf GSM8K} & \multirow{2}{*}{\bf ProofWriter} & \multirow{2}{*}{\bf FOLIO} & \multirow{2}{*}{\bf LogiQA} & \multirow{2}{*}{\bf Avg.} \\
        & & (6 digits) & (9 digits) & (2 digits) & (3 digits) &  & & &  \\
        \hline
        \multirow{2}{*}{Llama2-7B-Chat} & Direct & \textbf{0.618} * & \textbf{0.108} * & \textbf{0.242} * & \textbf{0.008} & 0.060 & 0.382 & 0.392 & \textbf{0.390} & \textbf{0.275} \\
        & CoT & 0.136 & 0.016 & 0.080 & 0.002 & \textbf{0.270} * & \textbf{0.398} & \textbf{0.480} & 0.372 & 0.219 \\
        \hline
        \multirow{2}{*}{Llama2-70B-Chat} & Direct & \textbf{0.826} * & \textbf{0.552} * & \textbf{0.488} * & \textbf{0.044} & 0.166 & \textbf{0.527} & 0.495 & \textbf{0.562} & \textbf{0.458} \\
        & CoT & 0.592 & 0.452 & 0.146 & 0.032 & \textbf{0.562} * & 0.523 & \textbf{0.500} & 0.548 & 0.419 \\
        \hline        
        \multirow{2}{*}{GPT-3.5-Turbo} & Direct & \textbf{0.962} * & \textbf{0.958} * & \textbf{1.000} * & \textbf{0.542} * & 0.300 & 0.277 & 0.505 & \textbf{0.548} * & \textbf{0.637} \\
        & CoT & 0.674 & 0.372 & 0.848 & 0.450 & \textbf{0.748} * & \textbf{0.518} * & \textbf{0.574} & 0.465 & 0.581 \\
        \hline
        \multirow{2}{*}{GPT-4} & Direct & 0.996 & 0.986 & 0.964 & 0.488 & 0.496 & 0.552 & \textbf{0.701} & \textbf{0.702} & 0.736 \\
        & CoT & \textbf{0.998} & \textbf{0.990} & \textbf{0.990} & \textbf{0.816} * & \textbf{0.946} * & \textbf{0.708} * & 0.686 & 0.688 & \textbf{0.853} \\
        \hline        
    \end{tabular}
    \caption{LLMs with CoT show mixed results, improving accuracy in some tasks while reducing it in others. The asterisk `*' indicates a statistical significance of $p$-value $< 0.01$ in McNemar's test. All experiments are in zero-shot settings.}
    \label{tab:direct_vs_cot.result}
\end{table*}

\begin{table*}[t]
    \centering\scriptsize
    \begin{tabular}{cccc@{\hspace{4pt}}ccccc}
        \hline
        \multirow{2}{*}{\bf LLM} & \multirow{2}{*}{{\bf CoT$\rightarrow$Answer}} & \multirow{2}{*}{\bf Error} & \multicolumn{2}{c}{\bf Simple Task} & \multicolumn{4}{c}{\bf Complex Task} \\
        \cmidrule(lr){4-5} \cmidrule(lr){6-9}
         &  &  & {{\bf Add.}(6 digits)} & {{\bf Mult.}(3 digits)} & {{\bf GSM8K}} & {{\bf ProofWriter}} & {{\bf FOLIO}} & {{\bf LogiQA}} \\
        \hline
        \multirow{4}{*}{GPT-3.5-Turbo} & \cmark $\rightarrow$ \cmark & - & 0.026 & 0.446 & 0.760 & 0.260 & 0.455 & 0.330 \\
        & \red{\it \cmark $\rightarrow$ \xmark} & \it \red{type 1} & \it 0.000 & \it \red{0.440} & \it 0.000 & \it 0.000  & \it 0.000 & \it \red{0.010} \\
        & \red{\it \xmark $\rightarrow$ \cmark} & \it \red{type 2} & \it \red{0.648} & \it \red{0.004} & \it 0.000 & \it \red{0.280} & \it \red{0.125} & \it \red{0.115} \\
        & \xmark $\rightarrow$ \xmark & type 3 & 0.326 & 0.110 & 0.240 & 0.460 & 0.420 & 0.545 \\
        \cline{2-9}
        & consistency error & type 1,2 & 0.648 & 0.444 & 0.000 & 0.280 & 0.125 & 0.125  \\
        \hline
        \multirow{4}{*}{GPT-4} & \cmark $\rightarrow$ \cmark & - & 0.254 & 0.804 & 0.945 & 0.640 & 0.630 & 0.620 \\
        & \red{\it \cmark $\rightarrow$ \xmark} & \red{\it type 1} & \it 0.000 & \it \red{0.166} & \it 0.000 & \it 0.000 & \it 0.000 & \it 0.000 \\
        & \red{\it \xmark $\rightarrow$ \cmark} & \red{\it type 2} & \it \red{0.744} & \it \red{0.012} & \it 0.000 & \it \red{0.060} & \it \red{0.070} & \it \red{0.010} \\
        & \xmark $\rightarrow$ \xmark & type 3 & 0.002 & 0.018 & 0.055 & 0.300 & 0.300 & 0.370 \\
        \cline{2-9}
        & consistency error & type 1,2 & 0.744 & 0.178 & 0.000 & 0.060 & 0.070 & 0.010  \\
        \hline        
    \end{tabular}
    \caption{\emph{Consistency errors} produced by LLMs with CoT, where the positive error rates are highlighted in \red{\it red}.}
    \label{tab:direct_vs_cot.confusion}
\end{table*}

In the intervention experiment, we randomly sample an instruction corresponding to a role for replacement and then observe the changes in the output of the LLM. The prompt provided to GPT-4 to generate the instructions is as follows:

\noindent
\begin{tabular}{p{19em}}
    \vspace{4pt}
    \textbf{\small Intervene on Instructions} \\
    \midrule
    \parbox{\linewidth}{\small 
        Below is a prompt to instruct LLM to tackle a problem, initially framed as a math teacher solving the issue. Your task is to rephrase the prompt with these adjustments: \\
        1. Request the LLM to assume the role of \{\{role\}\} while solving the problem. \\
        2. Alter the sentence structure to enhance differentiation. \\
        3. Add more interference to the prompt and make the prompt more different. \\
        4. Retain any specified output format requirements unchanged. \\
        5. Output the paraphrased prompt directly, do not incorporate other information. \\
        \# Prompt \\
        \{\{prompt\}\} \\
        \\
        \# Paraphrased prompt:
        }  \\
\end{tabular}

\section{CoT and Its Effectiveness}
\label{app:direct_vs_cot}
We conduct an empirical examination of CoT in six tasks in terms of task accuracy and reasoning errors. The results show that CoT does not consistently improve task performance, with instances where incorrect CoTs lead to correct answers and vice versa, indicating a potential spurious correlation between the CoTs and the answers.

\subsection{Variable Effects of CoT}
We assess the effectiveness of CoT by contrasting it with direct answering that does not involve step-by-step reasoning, as shown in Table~\ref{tab:direct_vs_cot.result}. The main observations are as follows.

\paragraph{CoT impairs performance in basic arithmetic tasks.}
LLMs have been found to successfully pass college entrance level exams \citep{openai2023gpt4}, which are considerably more challenging than primary school arithmetic from a human point of view. However, as the figure illustrates, CoT in basic arithmetic tasks reveals relatively low accuracies (below $0.55$) on GPT-3.5-turbo. Conversely, direct answers in Addition achieve significantly higher accuracies (above $0.95$).

\paragraph{CoT enhances performance in complex reasoning tasks.}
In the GSM8K math word problem, CoT consistently improves performance compared to direct answering, highlighting the effectiveness of step-by-step reasoning in solving more complex math problems. 
In logical reasoning problems, the improvement by CoT is relatively minor, but still consistent across FOLIO and ProofWriter. However, CoT struggles with real-world logical problems such as LogiQA, indicating a discrepancy.

\subsection{Inconsistent Behaviors of CoT}
\label{app:cot_inconsistency}
CoT performs poorly in basic arithmetic calculations but excels in complex mathematical and logical reasoning tasks, contradicting our intuition. We go into this by evaluating the confusion matrices of the CoT steps and answers as in Table~\ref{tab:direct_vs_cot.confusion}, finding even more puzzling behaviors of CoT.

\paragraph{Incorrect CoTs result in correct answers.}
A considerable portion (over $60\%$) of Addition samples exhibit this unusual behavior, where the reasoning steps are incorrect but yield the correct answers. This pattern persists even with larger LLMs, where the proportion increases to $74\%$ in GPT-4, suggesting that the problem may not be solved simply by enlarging the model.

A notable proportion of logical reasoning samples also display this peculiar correlation. A manual review in ProofWriter reveals that $28\%$ of the responses from GPT-3.5-turbo have correct answers but incorrect CoTs, which include fabricated facts and faulty inferences. The same pattern is also observed in FOLIO and LogiQA, albeit at a lower frequency, implying that the phenomenon is common in different tasks.

\paragraph{Correct CoTs result in incorrect answers.}
A significant proportion of Multiplication samples generate incorrect answers with correct reasoning steps. In these instances, we find that the answer steps either fail to gather the intermediate results from the reasoning steps or fail to correctly sum the results. This type of error is seldom seen in other tasks like GSM8K and logical problems because the answer steps in these tasks are straightforward and thus the models are less likely to fail.

\begin{figure*}[h!]
\tcbset{
    enhanced,
    colback=red!5!white,
    boxrule=0.1pt,
    colframe=red!75!black,
    fonttitle=\bfseries
}
\begin{minipage}{0.62\linewidth}
\begin{minipage}{1\linewidth}
    \scriptsize
    \begin{tcolorbox}[title=Addition (from GPT-3.5-Turbo), lifted shadow={1mm}{-2mm}{3mm}{0.1mm}{black!50!white}]
        Please act as a math teacher and solve the addition problem in the given template. ...\newline
        \#\#\#\#\newline
        \# Question:\newline
        What is the sum of 625126 and 542611?\newline
        \# Reasoning:\newline
        Let's add the two numbers digit by digit.\newline
        1. The ones place: 6 + 1 = 7\newline
        2. The tens place: 2 + 1 = 3\newline
        3. The hundreds place: \red{5 + 6 = 11}. Write down the 1 and carry over the 10.\newline
        4. The thousands place: \red{2 + 4 + 1 (carried over) = 7}\newline
        5. The ten thousands place: \red{6 + 5 = 11}. Write down the 1 and carry over the 10.\newline
        6. The hundred thousands place: \red{5 + 2 + 1 (carried over) = 8}\newline
        7. The millions place: \red{6}\newline
        Answer:\newline
        Therefore, the final computed sum is \red{1163737}.\newline
        {\it (Both the process and final result are incorrect, the correct answer should be 1167737.)}
    \end{tcolorbox}
\end{minipage}
\vfill
\begin{minipage}{1\linewidth}
    \scriptsize
    \begin{tcolorbox}[title=Multiplication (from GPT-4), lifted shadow={1mm}{-2mm}{3mm}{0.1mm}{black!50!white}]
        Please act as a math teacher and solve the product problem in the given template. ...\newline
        \#\#\#\#\newline
        \# Question:\newline
        What is the product of 858 and 624?\newline
        \# Reasoning:\newline
        Let's think step by step. 624 has three digits, so that we can reason in three steps.\newline
        1. Multiply 858 by the ones place digit 4 of 624, which is 3432.\newline
        2. Multiply 858 by the tens place digit 2 of 624, which is 1716 after the result is multiplied by 10 (since 2 is in the tens place) which results in 17160.\newline
        3. Multiply 858 by the hundreds place digit 6 of 624, which is 5148 after the result is multiplied by 100 (since 6 is in the hundreds place) which results in 514800.\newline
        Answer:\newline
        Now, sum all the step results: 3432 + 17160 + 514800 = \red{533392}.\newline
        So, the final computed product is \red{533392}.\newline
        {\it (Error in the total calculation, the correct sum should be 535392.)}
    \end{tcolorbox}
\end{minipage}
\end{minipage}
\hfill
\begin{minipage}{0.38\linewidth}
    \scriptsize
    \begin{tcolorbox}[title=ProofWriter (from GPT-4), lifted shadow={1mm}{-2mm}{3mm}{0.1mm}{black!50!white}]
        Please act as a math teacher and reason step by step to solve the logical reasoning problem. \newline
        ...\newline
        \#\#\#\#\newline
        \# Context:\newline
        Anne is furry. Anne is kind. Anne is nice. Anne is young. Bob is big. Bob is furry. Charlie is nice. Charlie is young. Dave is cold. Dave is kind. Dave is young. All cold, big people are furry. All big people are rough. All kind, big people are rough. If someone is nice and rough then they are kind. If Charlie is kind then Charlie is big. If someone is rough and kind then they are cold. Nice people are big. All rough, furry people are cold. Cold, nice people are big. \newline
        \# Question:\newline
        Based on the above information, is the following statement true, false, or unknown? Charlie is furry.\newline
        \# Reasoning:\newline
        Let's break down the information given:\newline
        1. Charlie is nice and young.\newline
        2. \red{If Charlie is kind then Charlie is big.}\newline
        3. Nice people are big. So, Charlie is big.\newline
        4. \red{If someone is nice and rough then they are kind. So, Charlie is kind.}\newline
        5. \red{If someone is rough and kind then they are cold. So, Charlie is cold.}\newline
        6. All cold, big people are furry. So, Charlie is furry.\newline
        Therefore, based on the given information, Charlie is furry. \newline
        The correct option is: A) True.\newline
        {\it (Step 2 is mentioned but not actually used, step 4 and 5 miss their precondition of `Charlie is rough', which can be derived from `All big people are rough'. The right answer is A.)}
    \end{tcolorbox}
\end{minipage}
\caption{Three examples of CoT mistakes, where either the CoT is incorrect but the answer is correct, or the other way around. The \red{red} highlights the incorrect steps, with explanations provided at the end of each.}
\label{fig:intro_case}
\end{figure*}

\subsection{Error Analysis}
\label{app:error_analysis}

LLMs with CoT fail on Addition and Multiplication mainly for two reasons: digit collection error and calculation error. The collection error occurs when a reasoning step locates a wrong digit for the calculation, such as step 3 in the failure case of Addition in Figure \ref{fig:intro_case}. The hundreds place for 625126 and 542611 is 1 and 6 but the step locates to 5 and 6 thereby causing the error. The calculation error could happen in any step involve a digital calculation, like the answer in the failure case of Multiplication in the figure, where the summation of 3432 + 17160 + 514800 gives 533392 but it should be 535392.

In ProofWriter, the model may identify the correct rule for reasoning through superficial keyword matching rather than logical reasoning, as shown in Figure \ref{fig:intro_case}. Even worse, the CoT not only makes incorrect causal reasoning, but invents factual information that is not provided in the context, yet the invented reasoning still leads to the correct answers.

\section{Discussion}
\label{app:discussion}

\paragraph{What is the difference between Llama2 and Mistral series?}
The 7B models including Llama2-7B and Mistral-7B perform poorly on Addition and Multiplication tasks, where the accuracies are less than 0.2. Among these tasks, Llama2 performs better on Addition, while Mistral performs better on Multiplication. Generally, the implied SCM types for Mistral-7B-DPO surpass those for Llama2-7B-Chat.

\paragraph{Do ChatGPT and GPT-4 perform perfectly on GSM8K like human reasoners?}
While both ChatGPT and GPT-4 suggest the ideal type I SCM on GSM8K, Tables \ref{tab:cot_intervention.result_chatgpt} and \ref{tab:cot_intervention.result_gpt4} indicate that the Average Treatment Effect (ATE) is not perfectly zero. This implies that although ChatGPT and GPT-4 perform similarly to a human in reasoning, they are not perfect due to their statistical nature.


\paragraph{Future Directions.}
In this study, we focus on the basic CoT, leaving the analysis of other alternatives like tree-of-thought and graph-of-thought to the future. Furthermore, our analysis is on a coarse-grained reasoning process, where it is simplified into only three random variables. More detailed analysis of fine-grained structures can be a direction for future work. The causal structures in LLMs could potentially be enhanced during the training of LLMs, for example, using counterfactual examples \cite{mitrovic2020representation,wu2021polyjuice,yang-2021-exploring} or causal regulation \cite{veitch2021counterfactual}, which can be worth exploration in future work.

\section{Detailed Results}
\label{app:detailed_results}

More detailed results are presented in following tables.

\begin{table*}[h]
    \centering\scriptsize
    \renewcommand{\arraystretch}{1.1}
    \begin{tabular}{lcccccc}
        \hline
        \multirow{2}{*}{\bf Intervention} & \multicolumn{6}{c}{\bf GPT-4}  \\ 
        & {\bf Addition} & {\bf Multiplication} & {\bf GSM8K} & {\bf ProofWriter} & {\bf FOLIO} & {\bf LogicQA} \\
        \hline
        CoT & 0.998 & 0.816 & 0.946 & 0.708 & 0.686 & 0.688 \\
        \hline
        \multicolumn{7}{c}{\bf Test: If CoT causes Answer given constant Instruction?} \\
        \grayline
        Controlled (w/ default setting) & 1.000 & 0.858 & 0.948 & 0.708 & 0.686 & 0.685 \\
        \quad\it Treated (w/ golden CoT) & \it +0.000 & \it +0.062 * & \it +0.052 * & \it +0.285 * & \it - & \it - \\
        \quad\it Treated (w/ random CoT) & \it -0.006 & \it -0.818 * & \it -0.016 & \it -0.082 * & \it -0.049 & \it -0.002 \\
        \grayline
        CoT $\overset{?}{\longrightarrow}$ Answer & F & T & T & T & F & F \\
        \hline
        \multicolumn{7}{c}{\bf Test: If Instruction causes Answer given constant CoT?} \\
        \grayline
        Controlled (w/ default setting) & 1.000 & 0.858 & 0.948 & 0.708 & 0.686 & 0.685 \\
        \quad\it Treated (w/ random instruction) & \it -0.002 & \it -0.044  & \it -0.002 & \it +0.000 & \it +0.000 & \it +0.003 \\
        \quad\it Treated (w/ random bias) & \it -0.174 * & \it -0.010 & \it -0.002 & \it -0.005 & \it -0.015 & \it -0.010 \\
        \grayline
        Controlled (w/ golden CoT) & 1.000 & 0.920 & 1.000 & 0.993 & \it - & \it - \\
        \quad\it Treated (w/ random instruction) & \it -0.002 & \it -0.008 & \it +0.000 & \it -0.007 & \it - & \it - \\
        \quad\it Treated (w/ random bias) & \it -0.154 * & \it -0.026 & \it -0.006 & \it -0.060 * & \it - & \it - \\
        \grayline
        Instruction $\overset{?}{\longrightarrow}$ Answer & T & F & F & T & F & F  \\
        \hline
        {\bf Implied SCM Type} & II & I & I & III & IV & IV  \\
        \hline
    \end{tabular}
    \caption{\emph{Identification of causal structures} in tasks running on \emph{GPT-4}. The symbol `*' denotes the average treatment effect (ATE) which is significant with a p-value less than 0.01.}
    \label{tab:cot_intervention.result_gpt4}
\end{table*}

\begin{table*}[h]
    \centering\scriptsize
    \renewcommand{\arraystretch}{1.1}
    \begin{tabular}{lcccccc}
        \hline
        \multirow{2}{*}{\bf Intervention} & \multicolumn{6}{c}{\bf Llama2-7B-Chat}  \\ 
        & {\bf Addition} & {\bf Multiplication} & {\bf GSM8K} & {\bf ProofWriter} & {\bf FOLIO} & {\bf LogicQA} \\
        \hline
        CoT & 0.136 & 0.002 & 0.270 & 0.398 & 0.480 & 0.372 \\
        \hline
        \multicolumn{7}{c}{\bf Test: If CoT causes Answer given constant Instruction?} \\
        \grayline
        Controlled (w/ default setting) & 0.120 & 0.002 & 0.280 & 0.400 & 0.510 & 0.335 \\
        \quad\it Treated (w/ golden CoT) & \it +0.052 & \it +0.010 & \it +0.518 * & \it +0.138 * & \it - & \it - \\
        \quad\it Treated (w/ random CoT) & \it -0.014 & \it -0.002 & \it -0.164 * & \it -0.020 & \it -0.020 & \it +0.005 \\
        \grayline
        CoT $\overset{?}{\longrightarrow}$ Answer & F & F & T & T & F & F  \\
        \hline
        \multicolumn{7}{c}{\bf Test: If Instruction causes Answer given constant CoT?} \\
        \grayline
        Controlled (w/ default setting) & 0.120 & 0.002 & 0.280 & 0.400 & 0.510 & 0.335 \\
        \quad\it Treated (w/ random instruction) & \it -0.052 * & \it +0.000 & \it -0.022 & \it -0.023 & \it -0.064 & \it -0.012 \\
        \quad\it Treated (w/ random bias) & \it -0.116 * & \it +0.000 & \it -0.084 * & \it -0.283 * & \it -0.324 * & \it -0.100 * \\
        \grayline
        Controlled (w/ golden CoT) & 0.172 & 0.012 & 0.798 & 0.538 & \it - & \it - \\
        \quad\it Treated (w/ random instruction) & \it -0.076 * & \it +0.016 & \it +0.008 & \it -0.060 & \it - & \it - \\
        \quad\it Treated (w/ random bias) & \it -0.172 * & \it +0.000 & \it -0.070 * & \it -0.480 * & \it - & \it - \\
        \grayline
        Instruction $\overset{?}{\longrightarrow}$ Answer & T & F & T & T & T  & T  \\
        \hline
        {\bf Implied SCM Type} & {\bf II} & {\bf IV} & {\bf III} & {\bf III} & {\bf II} & {\bf II}  \\
        \hline
    \end{tabular}
    \caption{\emph{Identification of causal structures} in tasks running on \emph{Llama2-7B-Chat}. The symbol `*' denotes the average treatment effect (ATE) which is significant with a p-value less than 0.01.}
    \label{tab:cot_intervention.result_llama2_7b}
\end{table*}

\begin{table*}[h]
    \centering\scriptsize
    \renewcommand{\arraystretch}{1.1}
    \begin{tabular}{lcccccc}
        \hline
        \multirow{2}{*}{\bf Intervention} & \multicolumn{6}{c}{\bf Llama2-70B-Chat}  \\ 
        & {\bf Addition} & {\bf Multiplication} & {\bf GSM8K} & {\bf ProofWriter} & {\bf FOLIO} & {\bf LogicQA} \\
        \hline
        CoT & 0.592 & 0.032 & 0.562 & 0.523 & 0.500 & 0.548 \\
        \hline
        \multicolumn{7}{c}{\bf Test: If CoT causes Answer given constant Instruction?} \\
        \grayline
        Controlled (w/ default setting) & 0.458 & 0.016 & 0.552 & 0.525 & 0.510 & 0.543 \\
        \quad\it Treated (w/ golden CoT) & \it -0.126 * & \it +0.180 * & \it +0.356 * & \it +0.092 * & \it - & \it - \\
        \quad\it Treated (w/ random CoT) & \it -0.062 * & \it +0.012 & \it -0.032 & \it -0.025 * & \it -0.044 & \it +0.002 \\
        \grayline
        CoT $\overset{?}{\longrightarrow}$ Answer & T & T & T & T & F & F \\
        \hline
        \multicolumn{7}{c}{\bf Test: If Instruction causes Answer given constant CoT?} \\
        \grayline
        Controlled (w/ default setting) & 0.458 & 0.016 & 0.552 & 0.525 & 0.510 & 0.543 \\
        \quad\it Treated (w/ random instruction) & \it -0.018 & \it +0.000 & \it +0.002 & \it -0.005 & \it +0.005 & \it -0.003 \\
        \quad\it Treated (w/ random bias)  & \it -0.036 & \it +0.018 & \it -0.024 & \it -0.008 & \it +0.000 & \it +0.000 \\
        \grayline
        Controlled (w/ golden CoT) & 0.332 & 0.196 & 0.908 & 0.617 & \it - & \it - \\
        \quad\it Treated (w/ random instruction) & \it +0.054 & \it -0.096 * & \it -0.048 * & \it -0.037 & \it - & \it - \\
        \quad\it Treated (w/ random bias) & \it -0.304 * & \it -0.052 * & \it -0.224 * & \it -0.208 * & \it - & \it - \\
        \grayline
        Instruction $\overset{?}{\longrightarrow}$ Answer & T & T & T & T & F & F  \\
        \hline
        {\bf Implied SCM Type} & {\bf III} & {\bf III} & {\bf III} & {\bf III} & {\bf IV} & {\bf IV}  \\
        \hline
    \end{tabular}
    \caption{\emph{Identification of causal structures} in tasks running on \emph{Llama2-70B-Chat}. The symbol `*' denotes the average treatment effect (ATE) which is significant with a p-value less than 0.01.}
    \label{tab:cot_intervention.result_llama2_70b}
\end{table*}

\begin{table*}[h!]
    \centering\scriptsize
    \renewcommand{\arraystretch}{1.1}
    \begin{tabular}{lccccccc}
        \hline
        \multirow{2}{*}{\bf Intervention} & \multicolumn{6}{c}{\bf GPT-3.5-Turbo (2-Shot)}   \\ 
        & {\bf Addition} & {\bf Multiplication} & {\bf GSM8K} & {\bf ProofWriter} & {\bf FOLIO} & {\bf LogicQA} & {\bf Avg. $|ATE|$} \\
        \hline
        CoT & 0.576 & 0.650 & 0.754 & 0.553 & 0.549 & 0.508 & \it - \\
        \hline
        \multicolumn{7}{c}{\bf Test: If CoT causes Answer given constant Instruction?} \\
        \grayline
        Controlled (w/ default setting) & 0.638 & 0.634 & 0.754 & 0.537 & 0.569 & 0.515& \it - \\
        \quad\it Treated (w/ golden CoT) & \it -0.128 * & \it +0.042 * & \it +0.246 * & \it +0.263 * & \it - & \it - & 0.170 \\
        \quad\it Treated (w/ random CoT) & \it +0.062 * & \it -0.634 * & \it -0.748 * & \it -0.245 * & \it -0.265 * & \it -0.043 * & 0.333 \\
        \grayline
        CoT $\overset{?}{\longrightarrow}$ Answer & T & T & T & T & T & T & 0.251  \\
        \hline
        \multicolumn{7}{c}{\bf Test: If Instruction causes Answer given constant CoT?} \\
        \grayline
        Controlled (w/ default setting) & 0.638 & 0.634 & 0.754 & 0.537 & 0.569 & 0.515 & \it - \\
        \quad\it Treated (w/ random instruction) & \it -0.214 * & \it -0.004 & \it +0.000 & \it +0.000 & \it -0.029 & \it -0.002 & 0.042 \\
        \quad\it Treated (w/ random bias) & \it -0.112 * & \it +0.066 * & \it +0.000 & \it -0.107 * & \it -0.088 * & \it -0.010 & 0.064 \\
        \grayline
        Controlled (w/ golden CoT) & 0.510 & 0.676 & 1.000 & 0.800 & \it - & \it - & \it - \\
        \quad\it Treated (w/ random instruction) & \it -0.078 * & \it +0.000 & \it +0.000 & \it +0.000 & \it - & \it - & 0.020 \\
        \quad\it Treated (w/ random bias) & \it -0.012 & \it +0.070 * & \it +0.000 & \it -0.362 * & \it - & \it - & 0.111 \\
                \grayline
        Instruction $\overset{?}{\longrightarrow}$ Answer & T & T & F & T & T  & F & 0.059 \\
        \hline
        {\bf Implied SCM Type} & {\bf III} & {\bf III} & {\bf I} & {\bf III} & {\bf III} & {\bf I} & \it - \\
        \hline
    \end{tabular}
    \caption{\emph{Identification of causal structures} in tasks running on \emph{GPT-3.5-Turbo with 2-Shot}.}
    \label{tab:cot_intervention.result_chatgpt_2shot}
\end{table*}

\begin{table*}[h!]
    \centering\scriptsize
    \renewcommand{\arraystretch}{1.1}
    \begin{tabular}{lccccccc}
        \hline
        \multirow{2}{*}{\bf Intervention} & \multicolumn{7}{c}{\bf GPT-3.5-Turbo (4-Shot)}   \\ 
        & {\bf Addition} & {\bf Multiplication} & {\bf GSM8K} & {\bf ProofWriter} & {\bf FOLIO} & {\bf LogicQA} & {\bf Avg. $|\text{ATE}|$} \\
        \hline
        CoT & 0.380 & 0.670 & 0.744 & 0.568 & 0.618 & 0.500 & \it - \\
        \hline
        \multicolumn{7}{c}{\bf Test: If CoT causes Answer given constant Instruction?} \\
        \grayline
        Controlled (w/ default setting) & 0.354 & 0.672 & 0.744 & 0.568 & 0.603 & 0.502 & \it - \\
        \quad\it Treated (w/ golden CoT) & \it +0.016 & \it +0.000 & \it +0.256 * & \it +0.270 * & \it - & \it - & 0.136 \\
        \quad\it Treated (w/ random CoT) & \it +0.012 & \it -0.672 * & \it -0.740 * & \it -0.232 * & \it -0.235 * & \it -0.022 & 0.319 \\
        \grayline
        CoT $\overset{?}{\longrightarrow}$ Answer & F & T & T & T & T & F & 0.227 \\
        \hline
        \multicolumn{7}{c}{\bf Test: If Instruction causes Answer given constant CoT?} \\
        \grayline
        Controlled (w/ default setting) & 0.354 & 0.672 & 0.744 & 0.568 & 0.603 & 0.502 & \it - \\
        \quad\it Treated (w/ random instruction) & \it +0.074 * & \it +0.010 & \it +0.000 & \it -0.007 & \it +0.005 & \it -0.002 & 0.016 \\
        \quad\it Treated (w/ random bias) & \it +0.008 & \it +0.058 * & \it +0.000 & \it -0.075 * & \it -0.123 * & \it -0.002 & 0.044 \\
        \grayline
        Controlled (w/ golden CoT) & 0.370 & 0.672 & 1.000 & 0.838 & \it - & \it - & \it - \\
        \quad\it Treated (w/ random instruction) & \it +0.040 & \it +0.002 & \it +0.000 & \it -0.040 & \it - & \it - & 0.021 \\
        \quad\it Treated (w/ random bias) & \it +0.052 & \it +0.066 * & \it +0.000 & \it -0.348 * & \it - & \it - & 0.117 \\
        \grayline
        Instruction $\overset{?}{\longrightarrow}$ Answer & T & T & F & T & T & F & 0.049 \\
        \hline
        {\bf Implied SCM Type} & {\bf II} & {\bf III} & {\bf I} & {\bf III} & {\bf III} & {\bf IV} & \it -  \\
        \hline
    \end{tabular}
    \caption{\emph{Identification of causal structures} in tasks running on \emph{GPT-3.5-Turbo with 4-Shot}.}
    \label{tab:cot_intervention.result_chatgpt_4shot}
\end{table*}

\begin{table*}[h!]
    \centering\scriptsize
    \renewcommand{\arraystretch}{1.1}
    \begin{tabular}{lccccccc}
        \hline
        \multirow{2}{*}{\bf Intervention} & \multicolumn{7}{c}{\bf GPT-3.5-Turbo (8-Shot)}   \\ 
        & {\bf Addition} & {\bf Multiplication} & {\bf GSM8K} & {\bf ProofWriter} & {\bf FOLIO} & {\bf LogicQA} & {\bf Avg. $|\text{ATE}|$} \\
        \hline
        CoT & 0.456 & 0.636 & 0.772 & 0.567 & 0.608 & 0.512 & \it - \\
        \hline
        \multicolumn{7}{c}{\bf Test: If CoT causes Answer given constant Instruction?} \\
        \grayline
        Controlled (w/ default setting) & 0.414 & 0.650 & 0.772 & 0.577 & 0.603 & 0.512 & \it - \\
        \quad\it Treated (w/ golden CoT) & \it +0.006 & \it +0.036 * & \it +0.228 * & \it +0.262 * & \it - & \it - & 0.133 \\
        \quad\it Treated (w/ random CoT) & \it +0.046 * & \it -0.648 * & \it -0.766 * & \it -0.272 * & \it -0.186 * & \it -0.035 * & 0.326 \\

        \grayline
        CoT $\overset{?}{\longrightarrow}$ Answer & T & T & T & T & T & T & 0.229 \\
        \hline
        \multicolumn{7}{c}{\bf Test: If Instruction causes Answer given constant CoT?} \\
        \grayline
        Controlled (w/ default setting) & 0.414 & 0.650 & 0.772 & 0.577 & 0.603 & 0.512 & \it - \\
        \quad\it Treated (w/ random instruction) & \it -0.016 & \it -0.016 & \it +0.000 & \it -0.040 * & \it -0.025 & \it +0.005 & 0.017 \\
        \quad\it Treated (w/ random bias) & \it +0.058 & \it +0.076 * & \it +0.000 & \it -0.190 * & \it -0.152 * & \it -0.008 & 0.081 \\
        \grayline
        Controlled (w/ golden CoT) & 0.420 & 0.686 & 1.000 & 0.838 & \it - & \it - & \it - \\
        \quad\it Treated (w/ random instruction) & \it +0.000 & \it +0.004 & \it +0.000 & \it -0.102 * & \it - & \it - & 0.027 \\
        \quad\it Treated (w/ random bias) & \it +0.076 * & \it +0.080 * & \it +0.000 & \it -0.393 * & \it - & \it - & 0.137 \\
        \grayline
        Instruction $\overset{?}{\longrightarrow}$ Answer & T & T & F & T & T & F & 0.065 \\
        \hline
        {\bf Implied SCM Type} & {\bf III} & {\bf III} & {\bf I} & {\bf III} & {\bf III} & {\bf I} & \it -  \\
        \hline
    \end{tabular}
    \caption{\emph{Identification of causal structures} in tasks running on \emph{GPT-3.5-Turbo with 8-Shot}.}
    \label{tab:cot_intervention.result_chatgpt_8shot}
\end{table*}

\begin{table*}[h!]
    \centering\scriptsize
    \renewcommand{\arraystretch}{1.1}
    \begin{tabular}{lccccccc}
        \hline
        \multirow{2}{*}{\bf Intervention} & \multicolumn{7}{c}{\bf Mistral-7B-Base (4-Shot)}   \\ 
        & {\bf Addition} & {\bf Multiplication} & {\bf GSM8K} & {\bf ProofWriter} & {\bf FOLIO} & {\bf LogicQA} & {\bf Avg. $|\text{ATE}|$} \\
        \hline
        CoT & 0.002 & 0.176 & 0.434 & 0.403 & 0.412 & 0.450 & \it - \\
        \hline
        \multicolumn{7}{c}{\bf Test: If CoT causes Answer given constant Instruction?} \\
        \grayline
        Controlled (w/ default setting) & 0.002 & 0.176 & 0.434 & 0.510 & 0.422 & 0.443 & \it - \\
        \quad\it Treated (w/ golden CoT) & \it +0.000 & \it +0.152 * & \it +0.566 * & \it +0.345 * & \it - & \it - & 0.266 \\
        \quad\it Treated (w/ random CoT) & \it +0.012 & \it -0.176 * & \it -0.426 * & \it -0.163 * & \it +0.034 & \it -0.040 * & 0.142 \\   
        \grayline
        CoT $\overset{?}{\longrightarrow}$ Answer & F & T & T & T & F & T & 0.204 \\
        \hline
        \multicolumn{7}{c}{\bf Test: If Instruction causes Answer given constant CoT?} \\
        \grayline
        Controlled (w/ default setting) & 0.002 & 0.176 & 0.434 & 0.510 & 0.422 & 0.443 & \it - \\
        \quad\it Treated (w/ random instruction) & \it -0.002 & \it -0.042 * & \it +0.000 & \it -0.135 * & \it -0.074 & \it -0.002 & 0.043 \\
        \quad\it Treated (w/ random bias) & \it -0.002 & \it -0.012 & \it -0.010 & \it -0.448 * & \it -0.103 * & \it +0.000 & 0.096 \\
        \grayline
        Controlled (w/ golden CoT) & 0.002 & 0.328 & 1.000 & 0.855 & \it - & \it - & \it - \\
        \quad\it Treated (w/ random instruction) & \it -0.002 & \it -0.090 * & \it -0.026 * & \it -0.343 * & \it - & \it - & 0.115 \\
        \quad\it Treated (w/ random bias) & \it +0.016 & \it -0.014 & \it -0.012 & \it -0.717 * & \it - & \it - & 0.190 \\
        \grayline
        Instruction $\overset{?}{\longrightarrow}$ Answer & F & T & T & T & T & F & 0.111 \\
        \hline
        {\bf Implied SCM Type} & {\bf IV} & {\bf III} & {\bf III} & {\bf III} & {\bf II} & {\bf I} & \it -  \\
        \hline
    \end{tabular}
    \caption{\emph{Identification of causal structures} in tasks running on \emph{Mistral-7B-Base}.}
    \label{tab:cot_intervention.result_mistral_base}
\end{table*}

\begin{table*}[h!]
    \centering\scriptsize
    \renewcommand{\arraystretch}{1.1}
    \begin{tabular}{lccccccc}
        \hline
        \multirow{2}{*}{\bf Intervention} & \multicolumn{7}{c}{\bf Mistral-7B-SFT (4-Shot)}   \\ 
        & {\bf Addition} & {\bf Multiplication} & {\bf GSM8K} & {\bf ProofWriter} & {\bf FOLIO} & {\bf LogicQA} & {\bf Avg. $|\text{ATE}|$} \\
        \hline
        CoT & 0.008 & 0.046 & 0.492 & 0.375 & 0.490 & 0.388 & \it - \\
        \hline
        \multicolumn{7}{c}{\bf Test: If CoT causes Answer given constant Instruction?} \\
        \grayline
        Controlled (w/ default setting) & 0.030 & 0.052 & 0.484 & 0.377 & 0.240 & 0.460 & \it - \\
        \quad\it Treated (w/ golden CoT) & \it -0.002 & \it +0.134 * & \it +0.510 * & \it +0.378 * & \it - & \it - & 0.256 \\
        \quad\it Treated (w/ random CoT) & \it +0.014 & \it -0.048 * & \it -0.470 * & \it +0.003 & \it +0.054 & \it -0.023 & 0.102 \\
        \grayline
        CoT $\overset{?}{\longrightarrow}$ Answer & F & T & T & T & F & F & 0.179 \\
        \hline
        \multicolumn{7}{c}{\bf Test: If Instruction causes Answer given constant CoT?} \\
        \grayline
        Controlled (w/ default setting) & 0.030 & 0.052 & 0.484 & 0.377 & 0.240 & 0.460 & \it - \\
        \quad\it Treated (w/ random instruction) & \it -0.012 & \it -0.038 * & \it -0.088 * & \it -0.195 * & \it -0.088 & \it -0.295 * & 0.119 \\
        \quad\it Treated (w/ random bias) & \it -0.030 * & \it -0.030 * & \it -0.046 * & \it -0.088 * & \it -0.083 * & \it -0.105 * & 0.064 \\
        \grayline
        Controlled (w/ golden CoT) & 0.028 & 0.186 & 0.994 & 0.755 & \it - & \it - & \it - \\
        \quad\it Treated (w/ random instruction) & \it -0.016 & \it -0.124 * & \it -0.168 * & \it -0.575 * & \it - & \it - & 0.221 \\
        \quad\it Treated (w/ random bias) & \it -0.026 * & \it -0.096 * & \it -0.090 * & \it -0.665 * & \it - & \it - & 0.219 \\
        \grayline
        Instruction $\overset{?}{\longrightarrow}$ Answer & T & T & T & T & T & T & 0.156 \\
        \hline
        {\bf Implied SCM Type} & {\bf II} & {\bf III} & {\bf III} & {\bf III} & {\bf II} & {\bf II} & \it -  \\
        \hline
    \end{tabular}
    \caption{\emph{Identification of causal structures} in tasks running on \emph{Mistral-7B-SFT}.}
    \label{tab:cot_intervention.result_mistral_sft}
\end{table*}

\begin{table*}[h!]
    \centering\scriptsize
    \renewcommand{\arraystretch}{1.1}
    \begin{tabular}{lccccccc}
        \hline
        \multirow{2}{*}{\bf Intervention} & \multicolumn{7}{c}{\bf Mistral-7B-DPO (4-Shot)}   \\ 
        & {\bf Addition} & {\bf Multiplication} & {\bf GSM8K} & {\bf ProofWriter} & {\bf FOLIO} & {\bf LogicQA} & {\bf Avg. $|\text{ATE}|$} \\
        \hline
        CoT & 0.000 & 0.012 & 0.326 & 0.322 & 0.520 & 0.470 & \it - \\
        \hline
        \multicolumn{7}{c}{\bf Test: If CoT causes Answer given constant Instruction?} \\
        \grayline
        Controlled (w/ default setting) & 0.000 & 0.010 & 0.324 & 0.310 & 0.397 & 0.485 & \it - \\
        \quad\it Treated (w/ golden CoT) & \it +0.004 & \it +0.048 * & \it +0.522 * & \it +0.238 * & \it - & \it - & 0.203 \\
        \quad\it Treated (w/ random CoT) & \it +0.008 & \it -0.006 & \it -0.312 & \it +0.000 & \it -0.049 & \it -0.058 * & 0.072 \\
        \grayline
        CoT $\overset{?}{\longrightarrow}$ Answer & F & T & T & T & F & T & 0.138 \\
        \hline
        \multicolumn{7}{c}{\bf Test: If Instruction causes Answer given constant CoT?} \\
        \grayline
        Controlled (w/ default setting) & 0.000 & 0.010 & 0.324 & 0.310 & 0.397 & 0.485 & \it - \\
        \quad\it Treated (w/ random instruction) & \it +0.034 * & \it +0.002 & \it -0.066 * & \it -0.067 * & \it -0.049 & \it -0.090 * & 0.051 \\
        \quad\it Treated (w/ random bias) & \it +0.002 & \it -0.006 & \it -0.028 & \it -0.035 & \it -0.034 & \it -0.145 * & 0.042 \\
        \grayline
        Controlled (w/ golden CoT) & 0.004 & 0.058 & 0.846 & 0.548 & \it - & \it - & \it - \\
        \quad\it Treated (w/ random instruction) & \it +0.034 * & \it +0.002 & \it -0.162 * & \it -0.153 * & \it - & \it - & 0.088 \\
        \quad\it Treated (w/ random bias) & \it +0.018 & \it -0.010 & \it -0.112 * & \it -0.200 * & \it - & \it - & 0.085 \\
        \grayline
        Instruction $\overset{?}{\longrightarrow}$ Answer & T & F & T & T & F & T & 0.066 \\
        \hline
        {\bf Implied SCM Type} & {\bf II} & {\bf I} & {\bf III} & {\bf III} & {\bf IV} & {\bf III} & \it -  \\
        \hline
    \end{tabular}
    \caption{\emph{Identification of causal structures} in tasks running on \emph{Mistral-7B-DPO}.}
    \label{tab:cot_intervention.result_mistral_dpo}
\end{table*}

\end{document}